\documentclass{article}

\ifdefined\pdfobjcompresslevel
\fi

\usepackage{microtype}
\usepackage{graphicx}
\usepackage{subcaption}
\usepackage{booktabs} 
\usepackage{multirow}

\usepackage{enumerate}
\usepackage{wrapfig}
\usepackage{hyperref}

\usepackage[accepted]{icml2026}

\usepackage{amsmath}
\usepackage{amssymb}
\usepackage{mathtools}
\usepackage{amsthm}

\usepackage{enumitem}

\usepackage[capitalize,noabbrev]{cleveref}

\theoremstyle{plain}

\theoremstyle{definition}

\theoremstyle{remark}

\usepackage[disable,textsize=tiny]{todonotes}

\icmltitlerunning{Multimodal Fusion via Self-Consistent Task-Gradient Fields}

\begin{document}

\twocolumn[
  \icmltitle{Multimodal Fusion via Self-Consistent Task-Gradient Fields}

  \begin{icmlauthorlist}
    \icmlauthor{Jiayu Xiong}{hqu,hqu_uni}
    \icmlauthor{Jing Wang}{hqu,hqu_uni}
    \icmlauthor{Jun Xue}{whu}
    \icmlauthor{Wanlong Wang}{hqu,hqu_uni}
    \icmlauthor{Jianlong Kwan}{hqu}
    \icmlauthor{Xiaosen Lyu}{hqu,hqu_uni}
    \icmlauthor{Zhouqiang Jiang}{osaka}
  \end{icmlauthorlist}

  \icmlaffiliation{hqu_uni}{Xiamen Key Laboratory of Computer Vision and Pattern Recognition, Huaqiao University, Xiamen, Fujian, China}
  \icmlaffiliation{hqu}{Huaqiao University, Xiamen, Fujian, China}
  \icmlaffiliation{whu}{Wuhan University, Wuhan, Hubei, China}
  \icmlaffiliation{osaka}{Nakashima Lab, SANKEN, The University of Osaka, Osaka, Japan}

  \icmlcorrespondingauthor{Jing Wang}{wroaring@hqu.edu.cn}

  \icmlkeywords{Machine Learning, ICML}

  \vskip 0.3in
]



\printAffiliationsAndNotice{}  

\begin{abstract}
Multimodal learning aims to preserve as much task-related information as possible from different inputs. However, current fusion designs often distort the feedback loop to feature extractors. Aggressively merging modalities entangles their representations, making the feature extractors fragile to incomplete inputs. Meanwhile, attempting to separate features via auxiliary losses frequently introduces optimization conflicts that distract from the primary task. We propose the Self-Consistent Field Autoencoder (SCFAE) to provide a better path for task gradients. Our method follows the self-consistent field principle to balance task learning with feature organization, thereby minimizing mutual information. We use small autoencoders for each modality to keep information intact. The task loss acts as a driving force to select predictive features. The reconstruction loss acts as a constraint to separate these features into independent subspaces. These dual objectives operate through complementary feature subspaces, thereby mitigating optimization interference. We evaluate SCFAE on audio-visual-text, audio-visual, and image-video benchmarks. Results show that SCFAE handles missing data and unequal input sizes more robustly via a simple structure. Gradient analysis confirms that SCFAE avoids conflicts and maintains stable training dynamics.
\end{abstract}

\section{Introduction}
\label{sec:intro}

Humans understand the world through multiple senses: vision, hearing, and language. Multimodal learning combines these different types of data to build better AI models~\citep{Baltrusaitis2019Survey}. When combined properly, models become more accurate and generalize better~\citep{Lu2019ViLBERT,Zhao2024CSUR}. The key challenge is designing the fusion stage, which controls how information from different modalities flows into the final prediction~\citep{Kendall2018Uncertainty,Chen2018GradNorm}.

Most fusion methods focus on representation quality. They use complex mechanisms like cross-modal attention, disentanglement, or contrastive learning to combine features. These methods improve test accuracy~\citep{MDF}, isolate missing modality~\citep{DrFuse}, or emphasize cross-modal cues~\citep{AVoiD-DF}. However, they often overlook a fundamental requirement: \textit{task gradients must reach the feature extractors cleanly}. The fusion stage should function as both an information coupler and a gradient distributor. When fusion breaks the feedback loop, extractors fail to learn meaningful representations. Since fusion can't create information that was never extracted~\citep{MMEntro}, ultimately throttling overall performance and leaving the model highly vulnerable to any breach in data integrity, such as missing modalities~\citep{QMF,Missing,MCULoRA}.

\begin{figure*}
    \centering
    \includegraphics[width=1\linewidth]{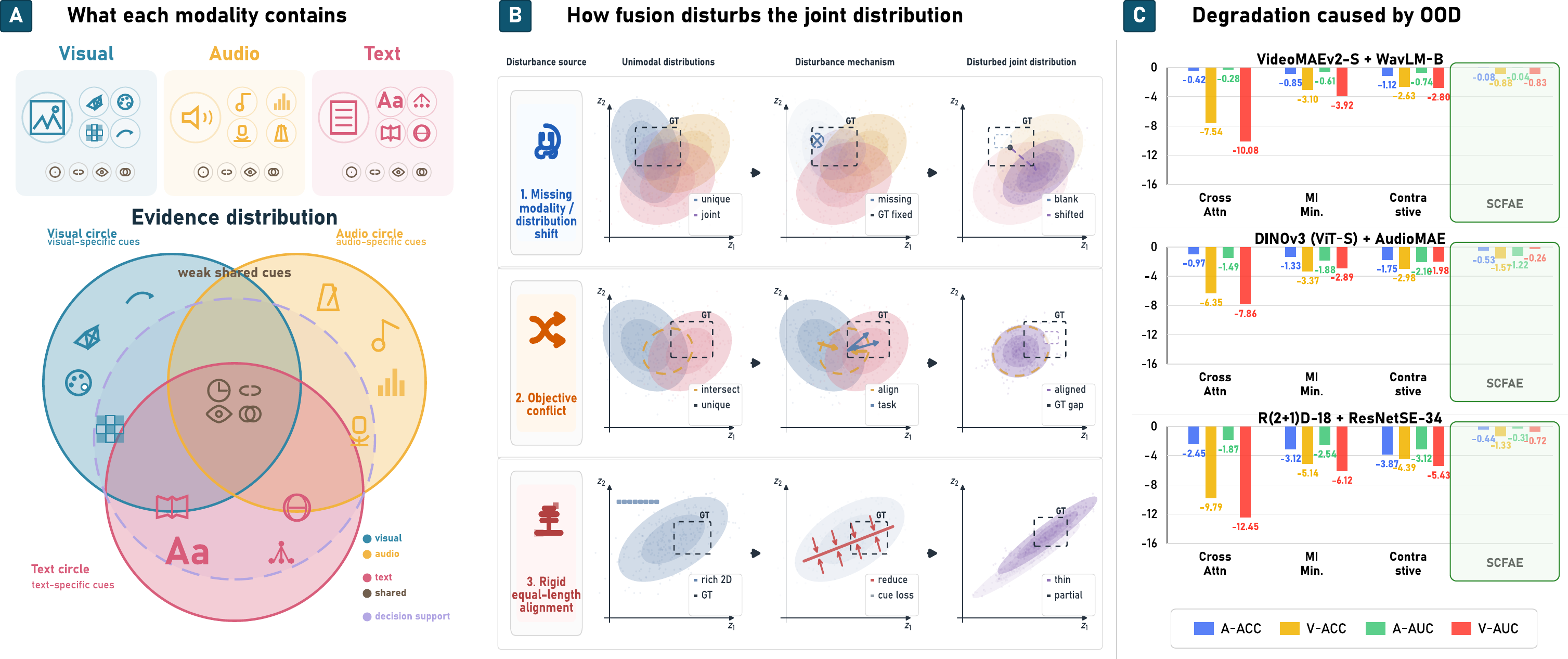}
    \caption{\textbf{Teaser of fusion-induced degradation.} \textbf{A:} Modalities contain specific cues, shared cues, and decision-supporting evidence. \textbf{B:} Missing inputs, auxiliary-objective conflict, and rigid alignment disturb this evidence geometry. \textbf{C:} The resulting degradation weakens unimodal extractors, while SCFAE preserves task-gradient access and organized subspaces. Complete results are reported in Tab.~\ref{tab:backbone_degradation}}
    \label{fig:intro_concept}

\end{figure*}

Specifically, Fig.~\ref{fig:intro_concept} illustrates how evidence geometry is damaged by distinct fusion failures that compromise the feature extractors. Coupled fusion aggressively mixes modalities before prediction~\citep{AVoiD-DF,MBT}, causing the extractors to become functionally codependent; this entanglement creates a fragility where the absence of one input collapses the shared gradient path, rendering the remaining extractors ineffective~\citep{QMF,MLA}. Decoupled fusion attempts to enforce independence via auxiliary losses (e.g., contrastive constraints~\citealt{MISA,DrFuse}), yet these secondary objectives often conflict with the primary task gradient, distracting the extractors from learning optimal predictive features~\citep{ContrastiveLearning,Missing}. Hybrid approaches risk inheriting both the fragility of entanglement and the optimization interference of auxiliary terms. Furthermore, to maintain computational tractability, these designs typically impose rigid dimensional constraints (e.g., forced alignment), compelling extractors to discard valuable temporal information when processing inputs of unequal lengths.

Hence, an ideal fusion framework must address these limitations by satisfying two fundamental requirements. First, it should preserve a clear optimization path, ensuring that task gradients provide direct feedback to the feature extractors~\citep{HARD1}. Second, it must isolate modal-specific features to minimize mutual information between modalities, which maintains robustness even when certain inputs are missing~\citep{LossEqualToMaxMI}. Successfully balancing these objectives prevents the typical trade-off between high task performance and model reliability.

Drawing on the Self-Consistent Field principle, we propose the Self-Consistent Field Autoencoder (SCFAE). We employ a dual-force optimization: the task loss guides feature selection, while the reconstruction loss enforces structural organization. This design captures the cross-modal synergy of coupled methods while maintaining the noise isolation of decoupled approaches. By shaping the feature topology rather than competing for semantic alignment, the reconstruction objective mitigates optimization interference. Our contributions are summarized as follows:
\begin{itemize}[leftmargin=*,itemsep=1pt,topsep=1pt]
\item Propose SCFAE to address gradient bottlenecks via the Self-Consistent Field principle. By treating task prediction and feature organization as distinct forces, it delivers clearer feedback to extractors than entangled designs.
\item Introduce a reconstruction-based strategy for information conservation. It organizes features into complementary subspaces that reduce redundancy while avoiding optimization conflicts from auxiliary constraints.
\item Evaluate SCFAE on benchmarks with missing modalities and unequal input lengths. Results demonstrate that our approach preserves stability and performance in settings where baseline methods degrade.
\end{itemize}

\begin{table}[t]
\centering
\footnotesize
\caption{Fusion-stage comparison. Align.: dimensional alignment required. Ref.: mutual-information--related objective/structure at fusion. Avail.: explicit design for missing modalities.}
\label{tab:feature_fusion}
\setlength{\tabcolsep}{1.5pt} 
\resizebox{\columnwidth}{!}{
\begin{tabular}{lcccc}
\toprule
Method & Align. & Ref. & Avail. & Modality \& Related Task \\
\midrule
CMMP~\citep{CLIP}            & \checkmark & \checkmark & --         & I and V, retrieval \\
METER~\citep{METER}          & \checkmark & \checkmark & --         & I and T, matching \\
APIVR~\citep{APIVR}          & \checkmark & \checkmark & --         & I and V, retrieval \\
MAP-IVR~\citep{MAP-IVR}      & $\times$   & \checkmark & --         & I and V, retrieval \\
DI-VTR~\citep{DI-VTR}        & \checkmark & \checkmark & --         & I and V, retrieval \\
AADML~\citep{AnchorDML}      & \checkmark & \checkmark & --         & I and V, retrieval \\
CMKT~\citep{CMKT}            & \checkmark & \checkmark & --         & I and V, retrieval \\
Coupled Mamba~\citep{CoupledMamba2024} & \checkmark & $\times$ & -- & Generic (I/V/A), retrieval/class. \\
ART-AVDF~\citep{WangHuang2024} & \checkmark & \checkmark & --       & A and V, deepfake \\
MACB-DF~\citep{MACBDF}       & \checkmark & \checkmark & --         & A and V, deepfake \\
CAFR~\citep{wang2025audio}   & \checkmark & \checkmark & --         & A and V, deepfake \\
MDS~\citep{MDS}              & \checkmark & \checkmark & --         & A and I, matching or classification \\
Emo-Foren~\citep{Emo}        & \checkmark & \checkmark & --         & A and V, matching or classification \\
VFD~\citep{VFD}              & \checkmark & \checkmark & --         & A and I, matching or classification \\
MISA~\citep{MISA}            & \checkmark & \checkmark & \checkmark & A V T, emotion classification \\
UAVM~\citep{UAVM}            & \checkmark & $\times$   & \checkmark & A and V, event matching \\
AVoiD-DF~\citep{AVoiD-DF}    & \checkmark & \checkmark & $\times$   & A and V, deepfake classification \\
DrFuse~\citep{DrFuse}        & \checkmark & \checkmark & \checkmark & EHR and CXR, classification \\
Perceiver~\citep{Perceiver}  & $\times$   & $\times$   & $\times$   & Generic \\
MCTN~\citep{MCTN}            & \checkmark & \checkmark & \checkmark & A V T, translation/sentiment \\
MMIN~\citep{MMIN}            & \checkmark & \checkmark & \checkmark & A V T, missing imputation \\
GCNet~\citep{GCNet}          & \checkmark & \checkmark & \checkmark & A V T, graph completion \\
IMDer~\citep{IMDer}          & \checkmark & \checkmark & \checkmark & A V T, sentiment analysis \\
DiCMoR~\citep{DiCMoR}        & \checkmark & \checkmark & \checkmark & A V T, sentiment analysis \\
MoMKE~\citep{MoMKE}          & \checkmark & \checkmark & \checkmark & A V T, sentiment analysis \\
EUAR~\citep{EUAR}            & \checkmark & \checkmark & \checkmark & A V T, uncertainty routing \\
MCULoRA~\citep{MCULoRA}      & $\times$   & $\times$   & \checkmark & A V T, low-rank adaptation \\
SALM~\citep{SALM2025}        & \checkmark & $\times$   & \checkmark & A V T, low-rank adaptation \\
DLF~\citep{DLF2025}          & \checkmark & \checkmark & \checkmark & A V T, disentanglement \\
AdaMMS~\citep{AdaMMS2025}    & \checkmark & \checkmark & \checkmark & Generic, model merging \\
\bottomrule
\end{tabular}
}
\vskip -1em
\end{table}

\section{Related Work}

We refine the broad categorization of \textbf{Coupled} and \textbf{Decoupled} fusion introduced in Sec.~\ref{sec:intro} into three operational axes detailed in Tab.~\ref{tab:feature_fusion}: \textbf{Align.} (dimensional pre-alignment), \textbf{Ref.} (auxiliary refinement objectives), and \textbf{Avail.} (support for missing modalities).

\textbf{Coupled Fusion} typically mandates strict dimensional alignment (\textbf{Align.}), creating the "gradient bottleneck" described earlier. As shown in the top section of Tab.~\ref{tab:feature_fusion}, retrieval-oriented methods (e.g., CMMP, METER) rely on alignment to enforce tight cross-modal interaction. While they employ Refinement objectives (\textbf{Ref.}) like contrastive loss, their goal is usually to force alignment rather than robustness, often lacking explicit design for missing inputs (\textbf{Avail.} is largely absent). This alignment-first strategy couples encoder learning to fixed dimensional constraints, leaving features fragile when inputs vary.

\textbf{Decoupled Fusion} attempts to break these rigid constraints to improve robustness (\textbf{Avail.}), as seen in the bottom section of Tab.~\ref{tab:feature_fusion} (e.g., MISA, DrFuse). However, to achieve this availability, these methods heavily rely on complex Refinement objectives (\textbf{Ref.}), such as mutual information minimization or orthogonality constraints. As argued in our Introduction, these auxiliary objectives act as the source of "optimization conflict," diverting gradients from the primary task. By analyzing methods through these axes, we pinpoint where existing designs sacrifice optimization stability for robustness, or vice versa. Neither coupled nor decoupled fusion fully preserves clean gradient flow to each encoder while maintaining feature separation.

\vspace{-0.02cm}
\section{Preliminary}
This section serves as a preparatory foundation, introducing the symbolic system, the optimization objectives from an entropic perspective, and the underlying physical background. 

\textbf{Formulation.} Consider inputs with $M$ modalities, where $m \in \{1, 2, \dots, M\}$ represents different modalities. A dataset containing $N$ samples is examined. Let the inputs be $X = \{X_1, X_2, \ldots, X_N\}$, where each specific sample $i \in \{1, 2, \dots, N\}$ is represented as $X_i = \{X_i^{(1)}, X_i^{(2)}, \ldots, X_i^{(M)}\}$. The outputs are $Y = \{Y_1, Y_2, \ldots, Y_N\}$, where each pair $\{X_i, Y_i\}$ forms a sample pair. $X^{(m)}_i$ denotes the original sample of modality $m$, which has different shapes, while the shape of $Y_i$ depends on the specific dataset and downstream task. For each modality $m$, a specific feature extractor $f^{(m)}(\cdot,\theta^{(m)})$ with parameters $\theta^{(m)}$ is employed for feature extraction. The feature generated from feature extractor $f^{(m)}(X^{(m)}_i,\theta^{(m)})$ is denoted as $V^{(m)}_i$. The fused feature representation capturing the multimodal interactions of sample $i$ is denoted as $Z_{i}=\{Z^{(1)}_i,Z^{(2)}_i,\cdots,Z^{(M)}_i\}$. 

\vspace{-0.015cm}
\subsection{Foundation of Multimodal Fusion}
\label{sec:fusion_foundation}

A fundamental principle guides our design: \textit{fusion cannot create information that was never extracted}. The fusion stage operates on fixed feature representations from modality-specific encoders; it can reorder, filter, or combine them, but cannot generate novel task-relevant patterns absent in the encoder outputs~\citep{MMEntro,kawaguchi2023information}. This information conservation principle has critical implications for optimization.

When fusion distorts the gradient pathway--whether through aggressive entanglement or conflicting auxiliary objectives--encoders receive degraded feedback about what features to extract. The system degrades not because fusion performs poorly on given features, but because encoders never learn to produce the right features in the first place. This gradient pathology is especially severe under distribution shift: if encoders rely on spurious correlations during training, no amount of sophisticated fusion can recover~\citep{GradRelated,ContrLoss}.

\textbf{Formalization.} Let $f(X,\theta)=\{f^{(m)}(X^{(m)},\theta^{(m)})\}_{m=1}^M$ denote modality-specific extractors with parameters $\theta^{(m)}$, $g(\cdot,\theta^F)$ the fusion module, and $h(\cdot,\theta^C)$ the task head. The ideal multimodal objective minimizes task uncertainty while ensuring each encoder remains locally optimal:
\begin{equation}
\begin{aligned}
&\min_{\theta,\theta^{F},\theta^{C}}\; H\!\left(Y \,\middle|\, h\!\left(g\!\big[f(X,\theta),\theta^{F}\big],\theta^{C}\right)\right) \\
&\text{s.t.}\quad \forall m\in\{1,\dots,M\},\\
&\theta^{(m)} \in \arg\min_{\hat{\theta}} \; H\!\left(Y \,\middle|\, f^{(m)}\!\left(X^{(m)},\hat{\theta}\right)\right).
\end{aligned}
\label{eq:ideal_objective}
\end{equation}

The constraint ensures fusion does not degrade modality-specific predictiveness--each encoder should extract maximum task-relevant information from its input. The global objective then leverages cross-modal complementarity. This two-stage decomposition sidesteps the need for explicit mutual information estimation; see Appendix~\ref{append:grad} for derivations.

\textbf{Architectural Implication.} This principle mandates explicit separation between extraction and fusion. Methods that bypass modality-specific encoders (e.g., feeding raw inputs to unified transformers) conflate these stages, making it impossible to enforce Eq.~\eqref{eq:ideal_objective}. Our work focuses on the classical multimodal pipeline--extract, fuse, predict--where fusion design directly impacts encoder learning.

\begin{figure}[t]
    \centering
    \includegraphics[width=\linewidth]{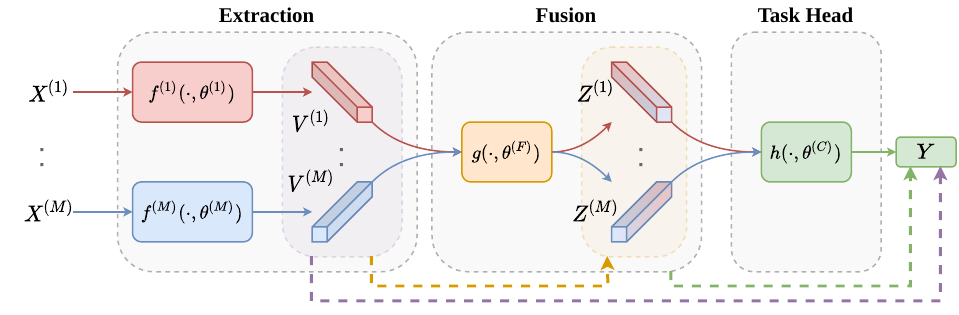}
    \caption{Stages of information entropy change. Where \(Z_i\) might be a set of vectors (\(\{Z_i^A, \dots,Z_i^M\}\)) or a vector, depending on the fusion method \(g(\cdot)\), and \(h(\cdot)\) stands for classifier.}
    \label{fig:eg}
    \vskip -1em
\end{figure}

\subsection{Self-Consistent Field for Gradient Coherence}
\label{sec:scf_principle}

\textbf{Problem.} The core challenge is to let task gradients reach encoders without distortion while preserving feature separation. Coupled fusion aggressively mixes modalities, making extractors fragile to missing inputs. Decoupled fusion instead relies on auxiliary separation losses, which can pull encoder parameters away from the primary task~\citep{ContrLoss,DrFuse}.

\textbf{Self-Consistent Field.} We seek a mechanism where multiple optimization forces coexist without interference. A \textit{self-consistent field} describes a feedback loop where the field governing a system depends on the system state itself. Formally, for a field $\phi$ and distribution $c$:
\begin{equation}
\phi = \mathcal{F}[c], \quad \frac{\partial c}{\partial t} = \mathcal{G}[\phi],
\end{equation}
where $\mathcal{F}, \mathcal{G}$ are functionals. Conceptually, the representation shapes its organizing signal, and that signal reshapes the representation.

\textbf{Poisson--Nernst--Planck (PNP).} PNP~\citep{PNP,Schuss2001DerivationPNP} is one illustrative SCF example where directional guidance and structural organization coexist. We use it only as intuition: multiple forces can remain coherent when coupled through a shared field rather than imposed as unrelated objectives.

\textbf{Conceptual Mapping to Multimodal Fusion.} We interpret multimodal features as analogous to particle distributions requiring organization:
\begin{itemize}[leftmargin=*,itemsep=1pt,topsep=1pt]
    \item \textbf{Unified Potential} $\Rightarrow$ A single scalar objective $\Phi$ from which all gradients derive, preventing conflicting updates.
    \item \textbf{Drift} $\Rightarrow$ Task-driven force guiding features toward task-relevant regions (analogous to an external electric field).
    \item \textbf{Diffusion} $\Rightarrow$ Information-preserving force maintaining feature separability and preventing collapse (analogous to concentration gradients).
    \item \textbf{Self-Consistency} $\Rightarrow$ The organizing force operates on features produced by the system itself.
\end{itemize}

The design principle is to achieve feature separation through complementary forces within a unified, self-consistent objective. 

\section{Method}
\label{sec:method}

\textbf{Architectural Scope.} SCFAE operates within the classical multimodal pipeline: $\text{Input} \xrightarrow{\text{Encoders}} \text{Features} \xrightarrow{\text{Fusion}} \text{Prediction}$. We assume modality-specific encoders $\{f^{(m)}\}$ are given or pre-trained, producing features $\{V^{(m)}_i \in \mathbb{R}^{l^{(m)}}\}$ of potentially different dimensions. Our contribution is the \textit{fusion block} $g(\cdot)$ that combines these features while satisfying the principles in Sec.~\ref{sec:fusion_foundation} and~\ref{sec:scf_principle}.
This design is orthogonal to end-to-end multimodal foundation models (e.g., ImageBind~\citep{girdhar2023imagebind}) that process raw inputs jointly. Our focus addresses scenarios where:
\begin{enumerate}[label=(\roman*),leftmargin=*,itemsep=1pt,topsep=1pt]
    \item Domain-specific encoders are preferred (e.g., medical imaging with specialized CNNs),
    \item Modalities arrive asynchronously or have distinct preprocessing pipelines,
    \item Interpretability requires tracking modality-specific contributions.
\end{enumerate}

\textbf{Architecture Overview.} Let \(V_i^{(m)}\in\mathbb{R}^{l^{(m)}}\) be the feature extractor output for modality \(m\), where \(i\) indexes the sample and \(l^{(m)}\) is the feature dimension. Define \(l^*=\min_m l^{(m)}\) as the minimum dimension across modalities. Our network uses four learned mappings per modality. Let \(\mathbf{P}\) denote a mapping network (SwiGLU + up proj. Linear, Appendix~\ref{append:expr}). The expansion map creates extra capacity before separation, so shared and specific evidence do not compete in the original feature space. The shared and specific projections then assign the two parts to task-facing and modality-facing pathways. The reconstruction map closes the loop by requiring the reconfigured representation to preserve the original encoder output. This keeps SCFAE lightweight: it changes only the fusion block and leaves the feature extractors untouched. The parameters $\theta^{\mathrm{SCFAE}}$ are:
\[
\Big\{\big(\mathbf{P}_{\mathrm{expand}}^{(m)},\mathbf{P}_{\mathrm{shared}}^{(m)},\mathbf{P}_{\mathrm{specific}}^{(m)},\mathbf{P}_{\mathrm{recon}}^{(m)}\big)\;|\;m=1,\dots,M\Big\}.
\]

\begin{figure*}[t]
    \centering
    \includegraphics[width=\linewidth]{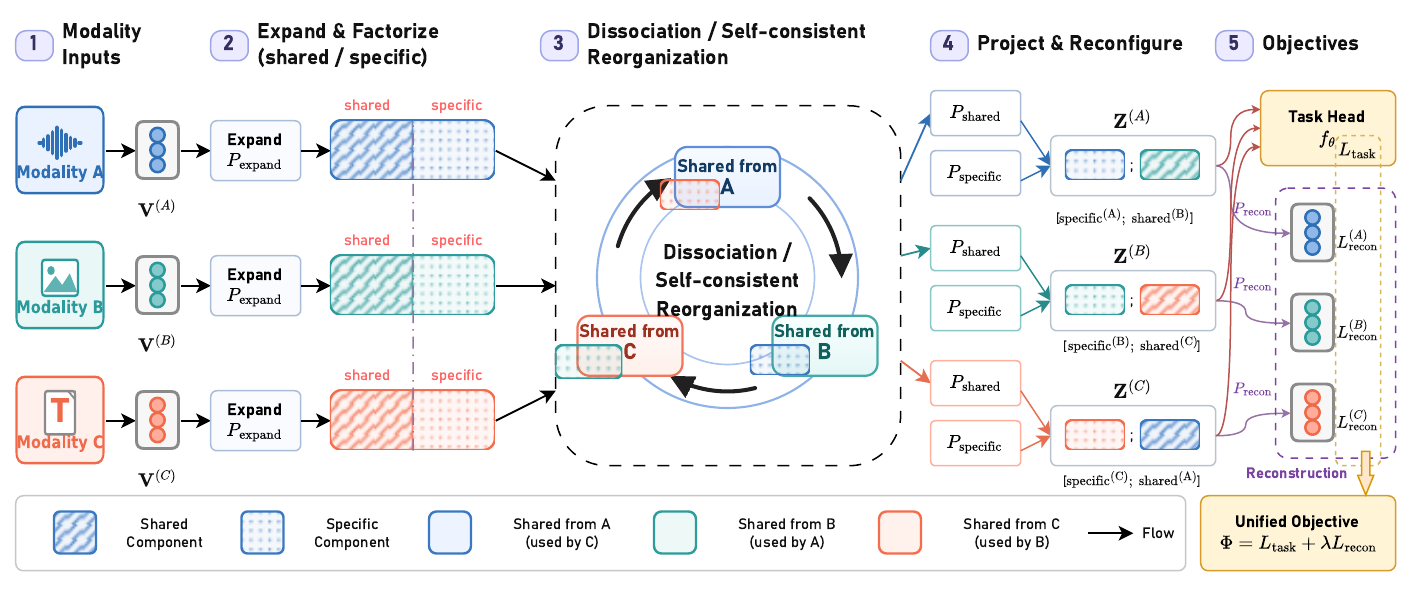}
    \caption{\textbf{SCFAE architecture.} Each modality is expanded and factorized into shared and specific components. The shared components are cyclically exchanged across modalities to form a self-consistent reference field, while each specific component stays with its own modality. After projection, the reconfigured features are optimized by a unified objective combining task prediction and reconstruction.}
    \label{fig:explain}
    \vskip -1em
\end{figure*}

\noindent\textbf{Expand.} Each feature \(V_i^{(m)}\) is mapped to a higher dimension \(n\,l^{(m)}\) and split at boundary \(b^{(m)}\):
\begin{equation}\label{eq:dis}
\begin{split}
\hat{Z}_i^{(m)} &= \mathbf{P}_{\mathrm{expand}}^{(m)} V_i^{(m)}, \\
\hat{Z}_{i,\mathrm{shared}}^{(m)} \in \mathbb{R}^{b^{(m)}}, & \quad \hat{Z}_{i,\mathrm{specific}}^{(m)} \in \mathbb{R}^{n l^{(m)}-b^{(m)}}.
\end{split}
\end{equation}
The expansion into higher dimensions provides geometric capacity for subsequent separation. Without this step, features would be forced to compete for the same representational space, preventing clean disentanglement. The boundary \(b^{(m)}\) pre-allocates how much capacity is reserved for shared versus specific information.

\noindent\textbf{Cross-modal reorganization.} Let \(k=(m+1)\bmod M\) denote the next modality. For modality \(m\), we replace its shared component with that from modality \(k\). This cyclic substitution is the self-consistent step: each modality is reorganized using another modality's shared evidence while retaining its own specific evidence. Both components are then projected to target sizes:
\begin{equation}\label{eq:con}
\begin{split}
Z_{i,\mathrm{shared}}^{(m)} &= \mathbf{P}_{\mathrm{shared}}^{(m)} \hat{Z}_{i,\mathrm{shared}}^{(k)}\in\mathbb{R}^{l^*}, \\
Z_{i,\mathrm{specific}}^{(m)} &= \mathbf{P}_{\mathrm{specific}}^{(m)} \hat{Z}_{i,\mathrm{specific}}^{(m)}\in\mathbb{R}^{l^{(m)}}.
\end{split}
\end{equation}
The reconfigured output concatenates both components: \(Z_i^{(m)}=[Z_{i,\mathrm{specific}}^{(m)}; Z_{i,\mathrm{shared}}^{(m)}]\in\mathbb{R}^{l^{(m)}+l^*}\). This serves as both the separated representation and the fusion output fed to the task head. Fig.~\ref{fig:explain} summarizes the full SCFAE path: expand each modality, split it into shared/specific parts, cyclically substitute the shared component, and jointly optimize task and reconstruction losses. Under optimization, the shared component should capture consistent cross-modal patterns (what all modalities agree on), while the specific component absorbs modality-unique noise and redundancy.

\noindent\textbf{Reconstruction.} A reconstruction mapping ensures no information is discarded during reorganization:
\begin{equation}\label{eq:masscon}
\mathcal{L}_{\mathrm{recon}}=\sum_{m=1}^M \text{Sim}(V_i^{(m)}, \mathbf{P}_{\mathrm{recon}}^{(m)} Z_i^{(m)}).
\end{equation}
where $\text{Sim}(\cdot, \cdot)$ is cosine similarity. This loss enforces the principle from Sec.~\ref{sec:fusion_foundation}: features can be reordered and split, but the total information content must remain recoverable. Without this constraint, the network could trivially achieve separation by discarding difficult modalities--exactly the failure mode we aim to prevent.

\noindent\textbf{Training Objective.} The complete loss function is:
\begin{equation}
\Phi = L_{\mathrm{task}} + \lambda L_{\mathrm{recon}},
\label{eq:total_loss}
\end{equation}
where \(\lambda\) balances task performance and information preservation. The task term trains the prediction path, while reconstruction prevents the split representation from discarding modality-specific evidence. Thus, $L_{\mathrm{task}}$ mainly shapes the shared subspace, and $L_{\mathrm{recon}}$ organizes the specific subspace.

\noindent\textbf{Connection to SCF.} The objective $\Phi$ echoes Sec.~\ref{sec:scf_principle}: all updates come from one unified objective, and the same representation is used for task prediction and feature organization. This forms a feedback loop between task-driven selection and information preservation, avoiding an auxiliary objective that competes with the primary task.

\section{Experiments}
\label{sec:expr}
This section first verifies that gradient mismatch under the setting in Sec.~\ref{sec:intro} is common in related multimodal approaches, and then evaluates SCFAE on multiple tasks.

\subsection{Experimental Setup}

\textbf{Hardware.} All experiments are conducted on a workstation equipped with an NVIDIA RTX 4090 GPU (24GB V-RAM) and an AMD Ryzen-9 5900X CPU (64GB RAM).

\textbf{Implementation Details.} We implement SCFAE using the PyTorch framework. To optimize memory usage, we employ the Apex library with 'O1' settings. We fix the random seed to '1' for all implementations to ensure reproducibility. Additional details are provided in the Appendix~\ref{append:expr}.

\begin{table}[t]
\centering
\caption{Summary of Datasets and Experimental Rationale.}
\label{tab:datasets}
\begin{center}
\resizebox{\columnwidth}{!}{
\begin{tabular}{l|l|l}
\toprule
\textbf{Dataset} & \textbf{Core Challenge} & \textbf{Physical Analogy} \\
\midrule
ActivityNet & Unequal-length inputs & \textbf{Conservation}: Keeps \\
(Retrieval) & (4096d vs 128d) & rich info without compression. \\
\midrule
FakeAVCeleb & Conflicting signals & \textbf{Drift}: Task loss guides \\
(Detection) & (Fake face vs real voice) & gradients to reliable data. \\
\midrule
CMU-MOSEI & Missing modalities & \textbf{Diffusion}: Features stay \\
(Sentiment) & (Randomly remove data) & organized in own subspaces. \\
\bottomrule
\end{tabular}
}
\end{center}
\vskip -1em
\end{table}

\textbf{Datasets.} 
We evaluate SCFAE on three benchmarks. Each dataset is chosen to test a specific challenge mentioned in our introduction. The relationship between the datasets and physical analogy is summarized in Tab.~\ref{tab:datasets}.

\noindent\textbf{Hyperparameters.} We have two main hyperparameters: the expansion factor \(n\) and the boundary ratio \(b\). Larger \(n\) increases representational capacity for separation; empirically, \(n \geq 2\) suffices (Sec.~\ref{sec:sensitivity}). The boundary \(b^{(m)}\) controls the shared-to-specific allocation; we use a uniform \(b^{(m)}=b=0.5\) across modalities as default.

\begin{table}[b!]
\vskip -1em
\centering
\caption{Comparison on ActivityNet (Image-Video retrieval). We report mAP@10/50/100.}
\label{tab:activitynet}
\resizebox{\linewidth}{!}{
\begin{tabular}{@{}p{4mm}|lccc@{}}
\toprule
& Method & mAP@10 & mAP@50 & mAP@100 \\
\midrule
\multirow{16}{*}{\rotatebox{90}{128-128}}
& CMMP~\citep{CLIP}        & 0.219 & 0.195 & 0.189 \\
& Perceiver~\citep{Perceiver}   & 0.271 & 0.253 & 0.240 \\
& METER~\citep{METER}       & 0.254 & 0.231 & 0.223 \\
& MISA~\citep{MISA}        & 0.263 & 0.238 & 0.229 \\
& MAP-IVR~\citep{MAP-IVR}     & 0.341 & 0.306 & 0.294 \\
& UAVM~\citep{UAVM}        & 0.340 & 0.308 & 0.297 \\
& DrFuse~\citep{DrFuse}      & 0.345 & 0.317 & 0.305 \\
& AADML~\citep{AnchorDML}       & 0.338 & 0.306 & 0.296 \\
& DI-VTR~\citep{DI-VTR}      & 0.347 & 0.320 & 0.306 \\
& CMKT~\citep{CMKT}        & 0.335 & 0.304 & 0.294 \\
& Coupled Mamba~\citep{CoupledMamba2024} & 0.328 & 0.301 & 0.290 \\
& DLF~\citep{DLF2025}       & 0.351 & 0.326 & 0.309 \\
& AdaMMS~\citep{AdaMMS2025}    & 0.358 & 0.339 & 0.315 \\
& \textbf{SCFAE} (Ours)     & \textbf{0.363} & \textbf{0.343} & \textbf{0.321} \\
\midrule
\multirow{10}{*}{\rotatebox{90}{4096-128}}
& APIVR~\citep{APIVR}       & 0.264 & 0.249 & 0.232 \\
& MAP-IVR~\citep{MAP-IVR}     & 0.349 & 0.322 & 0.311 \\
& MISA~\citep{MISA}        & 0.268 & 0.244 & 0.236 \\
& DrFuse~\citep{DrFuse}      & 0.352 & 0.325 & 0.313 \\
& Coupled Mamba~\citep{CoupledMamba2024} & 0.332 & 0.305 & 0.294 \\
& DLF~\citep{DLF2025}       & 0.355 & 0.331 & 0.314 \\
& AdaMMS~\citep{AdaMMS2025}    & 0.360 & 0.342 & 0.319 \\
& \textbf{SCFAE} (Ours)     & \textbf{0.367} & \textbf{0.349} & \textbf{0.326} \\
\bottomrule
\end{tabular}
}
\end{table}

\subsection{Handling Unequal-Length Inputs}
\label{subsec:activitynet}

\textbf{Setup.}
We study image-to-video retrieval on ActivityNet~\citep{ActivityNet}.
The goal is to retrieve videos that contain the same activity as a query image.
This task is hard because the modalities do not share the same feature shape, and videos include irrelevant parts.
Following prior ActivityNet retrieval works, we evaluate ranking quality with mean Average Precision at several standard cutoff ranks.%
~\citep{APIVR,MAP-IVR}
We test two input formats.
In the equal-length setting, we compress video features to match image feature length, so fusion can be done after alignment.
In the unequal-length setting, we keep the original long video features while leaving image features unchanged.
This setting tests whether a fusion block can use rich video signals without forcing early compression.

\textbf{Results.}
Tab.~\ref{tab:activitynet} reports the retrieval results under both input formats.
Methods requiring strict dimensional alignment struggle when input lengths differ.
SCFAE benefits from richer video signals because its self-consistent design avoids forced compression.
This confirms that the reconstruction constraint preserves information that alignment bottlenecks discard.
The trend holds across all cutoff ranks, showing that SCFAE turns unequal input sizes into an advantage rather than a limitation.

\textbf{Analysis.}
The main issue is how gradients and information move through the fusion stage.
Many coupled designs require early alignment.
This step can squeeze the video signal and make the task gradient less clear for the encoders.
Some decoupled designs rely on extra constraints that fight the task objective when modalities have very different capacity.
Our design uses reconstruction as a conservation constraint.
It keeps modality information recoverable, so the model does not need to discard video details to fit a shared size.
At the same time, the task loss decides what helps retrieval.
This matches our motivation: the fusion block should not block task gradients or add competing goals.
SCFAE turns unequal input lengths from a design constraint into a representational advantage, which is the gradient-preserving property our self-consistent field principle promises.

\subsection{Task-Driven Optimization under Conflict}
\label{subsec:fakeavceleb}

\textbf{Setup.}
We study a hard fusion case where the modalities disagree. In FakeAVCeleb~\citep{FAV}, a sample can contain a fake face but a real voice, or the opposite. This forces the fusion block to pick the right evidence, instead of averaging everything. We compare SCFAE with coupled fusion (Concat, UAVM) and decoupled fusion with extra constraints (MISA, DrFuse), plus a general fusion baseline (Perceiver) and audio-visual deepfake detectors that use inconsistency cues (e.g., AVoiD-DF).
We report Accuracy and ROC-AUC. Accuracy reflects performance at a fixed threshold.
ROC-AUC reflects ranking quality across thresholds, and is more informative when Accuracy is saturated.

\begin{table}[t]
\centering
\caption{Performance on the FakeAVCeleb dataset. 'A' and 'V' indicate using audio and visual modalities respectively (another modality is masked), 'AV' represents complete samples.}
\resizebox{\linewidth}{!}{
\begin{tabular}{lcccccc}
\toprule
Method & \multicolumn{3}{c}{ACC (\%)} & \multicolumn{3}{c}{AUC (\%)} \\
\cmidrule(lr){2-4}\cmidrule(lr){5-7}
 & A & V & AV & A & V & AV \\
\midrule
Concat    & 47.07 & 66.28 & 97.68 & 46.67 & 55.25 & 87.33 \\
MISA~\citep{MISA}        & 61.75 & 71.66 & 97.68 & 58.98 & 64.76 & 79.22 \\
DrFuse~\citep{DrFuse}      & 66.83 & 75.35 & 97.68 & 62.86 & 69.33 & 78.56 \\
Perceiver~\citep{Perceiver}   & 56.81 & 78.84 & 97.68 & 51.36 & 58.20 & 93.45 \\
ART-AVDF~\citep{WangHuang2024} & 66.30 & 80.70 & 96.40 & 68.90 & 84.30 & 98.20 \\
MACB-DF~\citep{MACBDF}  & -     & -     & 91.70 & -     & -     & 93.20 \\
CAFR~\citep{wang2025audio}     & -     & -     & 86.30 & -     & -     & 87.10 \\
AVoiD-DF~\citep{AVoiD-DF}    & 70.30 & 55.80 & 83.70 & 72.40 & 57.20 & 89.20 \\
VFD~\citep{VFD}         & -     & -     & 81.50 & -     & -     & 86.10 \\
Emo-Foren~\citep{Emo}   & -     & -     & 78.10 & -     & -     & 79.80 \\
MDS~\citep{MDS}         & -     & -     & 82.80 & -     & -     & 86.50 \\
Coupled Mamba~\citep{CoupledMamba2024} & 88.35 & 89.47 & 97.68 & 91.23 & 53.84 & 96.55 \\
SALM~\citep{SALM2025}    & 89.74 & 90.82 & 97.68 & 92.68 & 54.27 & 97.12 \\
DLF~\citep{DLF2025}      & 91.20 & 92.50 & 97.68 & 94.10 & 55.80 & 97.50 \\
AdaMMS~\citep{AdaMMS2025} & 93.45 & 93.10 & 97.68 & 96.30 & 56.10 & 98.25 \\
\midrule
\textbf{SCFAE} (Ours)  & \textbf{95.74} & \textbf{95.35} & 97.68 & \textbf{99.51} & 56.40 & \textbf{98.70} \\
\bottomrule
\end{tabular}
}
\label{tab:fakeavceleb}
\end{table}

\textbf{Results.}
Tab.~\ref{tab:fakeavceleb} shows two clear patterns. First, several methods reach similar Accuracy on full inputs, but their ROC-AUC varies widely. This gap confirms that accuracy alone masks fusion quality. Second, methods with auxiliary constraints can lose to simple fusion when modalities disagree. This suggests that a fixed separation rule is brittle under conflicting signals.

\begin{table*}[t]
\centering
\caption{Accuracy comparison under fixed missing protocol on CMU-MOSEI. Our method consistently outperforms baselines, especially when the dominant text modality is missing.}
\setlength{\tabcolsep}{10pt}
\resizebox{\textwidth}{!}{
\begin{tabular}{l c c c c c c c c}
\toprule
\multirow{2}{*}{Models} &
\multicolumn{7}{c}{Testing condition (ACC/F1, \%)} &
\multirow{2}{*}{$\{a,t,v\}$} \\
\cmidrule(lr){2-8}
 & $\{a\}$ & $\{t\}$ & $\{v\}$ & $\{a,v\}$ & $\{a,t\}$ & $\{t,v\}$ & Avg. & \\
\midrule
MCTN~\cite{MCTN}      & 62.7/54.5 & 82.6/82.8 & 62.6/57.1 & 63.7/62.7 & 83.5/83.3 & 83.2/83.2 & 73.1/70.6 & 84.2/84.2 \\
MMIN~\cite{MMIN}      & 58.9/59.5 & 82.3/82.4 & 59.3/60.0 & 63.5/61.9 & 83.7/83.3 & 83.8/83.4 & 71.9/71.8 & 84.3/84.2 \\
GCNet~\cite{GCNet}    & 60.2/60.3 & 83.0/83.2 & 61.9/61.6 & 64.1/57.2 & 84.3/84.4 & 84.3/84.4 & 73.1/72.8 & 85.2/85.1 \\
IMDer~\cite{IMDer}    & 63.8/60.6 & 84.5/84.5 & 63.9/63.6 & 64.9/63.5 & 85.1/85.1 & 85.0/85.0 & 76.0/75.3 & 85.1/85.1 \\
DiCMoR~\cite{DiCMoR}  & 62.9/60.4 & 84.3/84.4 & 63.6/63.6 & 65.2/64.4 & 85.0/84.9 & 85.0/84.9 & 75.9/75.4 & 85.1/85.1 \\
MoMKE~\cite{MoMKE}    & 65.9/65.5 & 80.8/80.7 & 64.9/64.8 & 65.9/65.5 & 86.0/85.9 & 84.4/84.3 & 76.4/76.2 & 86.7/86.6 \\
EUAR~\cite{EUAR}      & 64.5/60.7 & 85.3/85.2 & 66.4/65.3 & 66.5/65.4 & 85.6/85.1 & 86.0/86.0 & 77.3/76.3 & 86.6/86.4 \\
Coupled Mamba~\cite{CoupledMamba2024}   & 67.8/69.4 & 86.5/86.7 & 68.9/70.5 & 70.3/71.7 & 86.8/87.0 & 86.6/86.8 & 79.0/79.8 & 86.9/87.1 \\
MCULoRA~\cite{MCULoRA}& 68.5/70.2 & 86.9/87.0 & 69.6/71.1 & 71.0/72.4 & 87.2/87.3 & 87.0/87.1 & 79.6/80.3 & 87.2/87.3 \\
SALM~\cite{SALM2025}           & 68.3/69.9 & 86.8/87.0 & 69.4/71.0 & 70.8/72.2 & 87.1/87.3 & 86.9/87.1 & 79.4/80.2 & 87.2/87.4 \\
DLF~\cite{DLF2025}    & 66.8/68.1 & 87.1/87.2 & 68.5/69.9 & 69.5/70.8 & 87.5/87.6 & 87.3/87.5 & 79.9/80.5 & 87.6/87.8 \\
AdaMMS~\cite{AdaMMS2025} & 67.2/69.0 & 86.9/87.1 & 69.8/71.5 & 71.2/72.5 & 87.3/87.5 & 87.2/87.4 & 80.1/80.8 & 87.4/87.7 \\
\midrule
SCFAE                   & \textbf{69.3}/\textbf{71.0} & \textbf{87.3}/\textbf{87.4} & \textbf{70.4}/\textbf{72.0} & \textbf{72.0}/\textbf{73.3} & \textbf{87.7}/\textbf{87.9} & \textbf{87.6}/\textbf{87.8} & \textbf{80.3}/\textbf{81.1} & \textbf{87.8}/\textbf{88.0} \\
\bottomrule
\end{tabular}
\label{tab:mosei}
}
\vskip -0.5em
\end{table*} 

\begin{table}[t]
\centering
\caption{Impact of Fusion Strategies on Feature Extractor Degradation (FakeAVCeleb). Row 0 shows the absolute performance of standalone unimodal training. Subsequent rows show the performance gap (\textbf{REMOVE} fusion, linear probing $\Delta$ of extractor's representation, lower decline is better). Negative values indicate that the joint training process damaged the feature extractor's specific representational power. \textbf{Bold} indicates the minimum degradation.}
\resizebox{\linewidth}{!}{
\begin{tabular}{l|cc|cc}
\toprule
\multirow{2}{*}{\textbf{Method}} & \multicolumn{2}{c|}{\textbf{Accuracy (ACC) \%}} & \multicolumn{2}{c}{\textbf{ROC-AUC \%}} \\
& Audio ($\Delta$) & Visual ($\Delta$) & Audio ($\Delta$) & Visual ($\Delta$) \\
\midrule
\multicolumn{5}{l}{\textit{\textbf{VideoMAE V2-Small~\citep{wang2023videomae} + WavLM-Base~\citep{chen2022wavlm}}}} \\
\midrule
0. Standalone Unimodal Baseline & 99.24 & 98.15 & 99.58 & 96.77 \\
\midrule
1. Cross-Attention (4-Layers) & -0.42 & -7.54 & -0.28 & -10.08 \\
2. Mutual Information Min. & -0.85 & -3.10 & -0.61 & -3.92 \\
3. Contrastive Constraints & -1.12 & -2.63 & -0.74 & -2.80 \\
4. \textbf{SCFAE (Ours)} & \textbf{-0.08} & \textbf{-0.88} & \textbf{-0.04} & \textbf{-0.83} \\
\midrule
\midrule
\multicolumn{5}{l}{\textit{\textbf{DINOv3 (ViT-S)~\citep{simeoni2025dinov3} + AudioMAE~\citep{huang2022masked}}}} \\
\midrule
0. Standalone Unimodal Baseline & 98.17 & 99.35 & 98.19 & 97.45 \\
\midrule
1. Cross-Attention (4-Layers) & -0.97 & -6.35 & -1.49 & -7.86 \\
2. Mutual Information Min. & -1.33 & -3.37 & -1.88 & -2.89 \\
3. Contrastive Constraints & -1.75 & -2.98 & -2.10 & -1.98 \\
4. \textbf{SCFAE (Ours)} & \textbf{-0.53} & \textbf{-1.57} & \textbf{-1.22} & \textbf{-0.26} \\
\midrule
\midrule
\multicolumn{5}{l}{\textit{\textbf{R(2+1)D-18~\citep{R2+1D} + ResNetSE-34~\citep{ResNet}} (End-to-End CNNs)}} \\
\midrule
0. Standalone Unimodal Baseline & 94.82 & 95.35 & 96.43 & 88.67 \\
\midrule
1. Cross-Attention (4-Layers) & -2.45 & -9.79 & -1.87 & -12.45 \\
2. Mutual Information Min. & -3.12 & -5.14 & -2.54 & -6.12 \\
3. Contrastive Constraints & -3.87 & -4.39 & -3.12 & -5.43 \\
4. \textbf{SCFAE (Ours)} & \textbf{-0.44} & \textbf{-1.33} & \textbf{-0.31} & \textbf{-0.72} \\
\bottomrule
\end{tabular}
}
\label{tab:backbone_degradation}
\vskip -1.8em
\end{table}

\textbf{Analysis.}
This experiment matches the problem in the introduction. Conflict is the case where gradients are most likely to collide. Auxiliary objectives try to enforce a stable geometry, even when the task needs the two modalities to disagree. This can weaken the task signal that should reach the encoders. SCFAE avoids this failure mode. The task loss decides which modality to trust for the label. The reconstruction term keeps information recoverable, so the model cannot win by deleting a hard modality. Fusion becomes a routing problem rather than a forced agreement. This shows that task-driven routing, not forced alignment, is the key to robust fusion under conflict.

\textbf{Extractor Quality.}
Tab.~\ref{tab:backbone_degradation} compares several fusion schemes by checking whether joint training weakens the two unimodal feature extractors. We employ a 1. cross-attention structure~\citep{AVoiD-DF} with most similar parameter counts; The network is consistent with SCFAE, but 2. The losses are consistent with~\citet{MISA} and ~\citet{DrFuse}. 3. The structure of the losses is consistent with~\citet{VFD} and is compared with SCFAE. The main pattern is consistent across all backbone settings. Architectures that mix modalities too early distort the modality-specific space, with more damage on the visual side. SCFAE preserves the unimodal encoders near their original quality. This aligns with our design goal. Task gradients coordinate the two streams through a self-consistent field without direct feature overriding. The learned representations remain transferable when each modality is probed independently. This robustness confirms that SCFAE protects encoder quality under joint training.

\subsection{Robustness to Missing Modalities}
\label{subsec:mosei}
\textbf{Setup.}
We assess how well the model handles missing data using the CMU-MOSEI~\citep{MOSEI} dataset. We train the model using all three modalities: audio, visual, and text. However, during testing, we remove one or two modalities to see if the model remains stable. We evaluate all seven possible input combinations. We report Accuracy and F1 score as our metrics.

\textbf{Results.}
Tab.~\ref{tab:mosei} reveals a clear trend. Text provides the strongest signal, so almost all methods perform well when text is present. Performance drops significantly for most baselines when they rely only on audio or visual inputs. SCFAE maintains stable performance even without the dominant modality. This confirms that the diffusion force keeps non-dominant features organized and task-ready regardless of input availability.

\textbf{Analysis.}
These results confirm our hypothesis about gradient flow. In standard methods, the text modality is too dominant. The strong text signal drives optimization, creating an easy shortcut that ignores audio-visual features. Their encoders do not receive enough useful feedback to learn robust patterns. Our method uses the "Diffusion" force from the Introduction. The reconstruction loss keeps audio and visual features organized and distinct in their own subspace, regardless of the text signal. When text input is removed during testing, those features remain high-quality and task-ready. This validates that the diffusion force protects non-dominant modalities from being suppressed during joint training.

\begin{table}[t]
    \centering
    \caption{Modality-order robustness of SCFAE on CMU-MOSEI. Grad. norm is the norm of the task gradient reaching the encoder; Neg. is the proportion of negative cosine similarity between task and reconstruction gradients.}
    \label{tab:mosei_order}
    \setlength{\tabcolsep}{4pt}
    \resizebox{\columnwidth}{!}{
    \begin{tabular}{lccccc}
    \toprule
    Order & $\{a,t,v\}$ & $\{a,v\}$ & Avg. & Grad. norm & Neg. \\
    \midrule
    $A{\rightarrow}V{\rightarrow}T$ & 87.8/88.0 & 72.0/73.3 & 80.3/81.1 & 2.08 & 0.08 \\
    $A{\rightarrow}T{\rightarrow}V$ & 87.7/87.9 & 71.8/73.1 & 80.1/80.9 & 2.03 & 0.09 \\
    $V{\rightarrow}A{\rightarrow}T$ & 87.9/88.0 & 72.0/73.3 & 80.3/81.0 & 2.05 & 0.08 \\
    \bottomrule
    \end{tabular}
    }
\end{table}

\textbf{Order robustness.}
Because SCFAE uses a cyclic shared-reference path, we test whether the tri-modal case is sensitive to modality order. Tab.~\ref{tab:mosei_order} shows that permuting the order produces negligible variation in both accuracy and gradient statistics. The cyclic path therefore behaves as a local shared reference rather than a fragile ordering artifact.

\begin{table}[t!]
    \centering
    \caption{Structural ablation on ActivityNet (128--128 setting). We report mAP@10/50/100.
    \textbf{Task}: task loss. \textbf{Rec.}: reconstruction loss. \textbf{Exp.}: expansion mechanism ($n{>}1$). Rec.: reconstruction loss (ours); Con.: contrastive loss (replacement).}
    \label{tab:ablation}
    \resizebox{1.0\linewidth}{!}{
    \begin{tabular}{lcccccc}
    \toprule
    Method & Task & Rec. & Exp. & mAP@10 & mAP@50 & mAP@100 \\
    \midrule
    Linear Proj. & \checkmark & -- & -- & 0.245 & 0.221 & 0.210 \\
    FFN (w/o Recon) & \checkmark & -- & \checkmark & 0.284 & 0.255 & 0.241 \\
    No Expansion ($n{=}1$) & \checkmark & \checkmark & -- & 0.335 & 0.312 & 0.298 \\
    Contrastive Replacement & \checkmark & Con. & \checkmark & 0.331 & 0.309 & 0.295 \\
    \midrule
    \textbf{SCFAE (Ours)} & \checkmark & \checkmark & \checkmark & \textbf{0.363} & \textbf{0.343} & \textbf{0.321} \\
    \bottomrule
    \end{tabular}
    }
    \vskip -1em
\end{table}

\begin{figure*}[t]
    \centering
    \includegraphics[width=\linewidth]{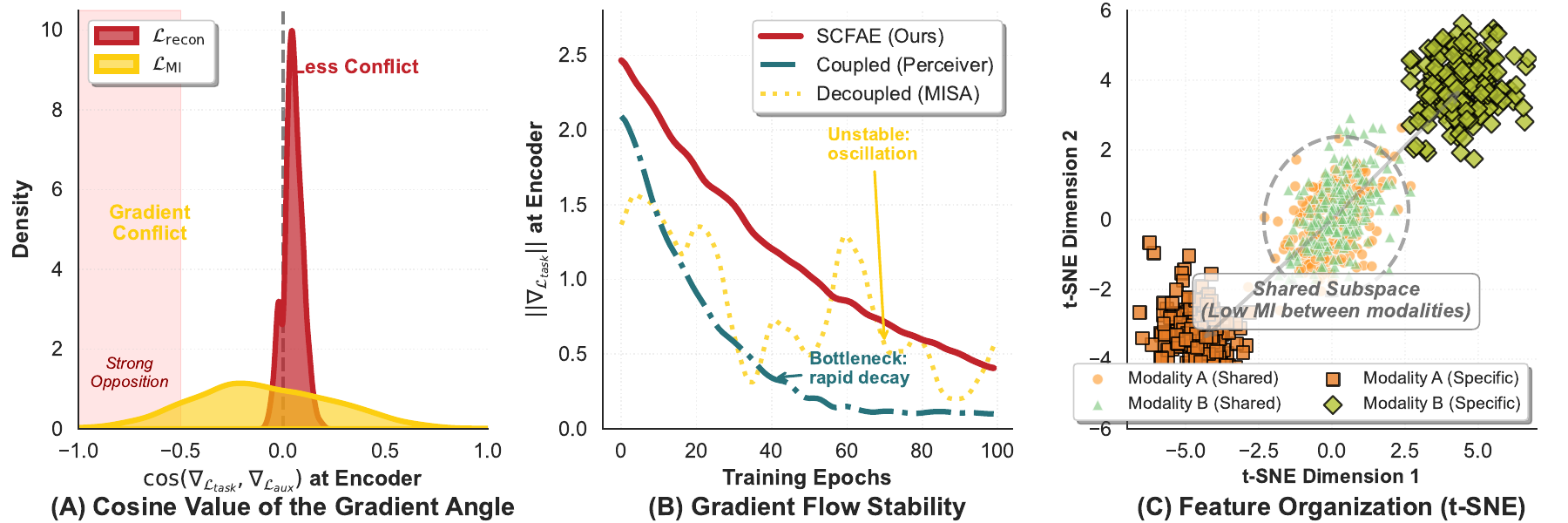}
    \caption{Visualization of gradient behavior and feature organization. \textbf{(A) Gradient cosine angle (FakeAVCeleb)}: We measure the angle between the task gradient and the auxiliary gradient. SCFAE (red) is centered at 0, indicating nearly orthogonal objectives that do not conflict. Baselines (yellow) show interference. \textbf{(B) Gradient Flow Stability (CMU-MOSEI)}: We track the magnitude of task gradients reaching the encoder. SCFAE maintains strong flow, while coupled fusion bottlenecks the signal. \textbf{(C) Feature Organization (FakeAVCeleb)}: t-SNE plot shows that SCFAE successfully pulls shared features (circles) together while pushing specific features (squares/diamonds) apart to a noise buffer zone.}
    \label{fig:grads}
    \vskip -1em
\end{figure*}

\subsection{Structural Ablation}
\label{sec:ablation}

We study how each SCFAE component affects ActivityNet image-to-video retrieval.
All variants keep the same feature dimensions for a fair comparison.
We use the shared representation $Z_{\text{shared}}$ for retrieval, and report results in Tab.~\ref{tab:ablation}.

\textbf{Expansion mechanism.}
Replacing the expansion-and-split design with a single linear projection degrades retrieval performance.
Using no expansion ($n{=}1$) is better than direct projection, but still underperforms the full model.
These results indicate that expansion is needed to separate shared and modality-specific information effectively.

\textbf{Reconstruction loss.}
Keeping the expansion structure but removing the reconstruction loss also hurts performance.
This suggests that the architecture alone is not enough.
The reconstruction loss encourages information preservation, which helps the specific component retain modality-unique signals while allowing the shared component to capture cross-modal patterns.

\textbf{Objective design.}
Replacing the reconstruction loss with a contrastive loss on shared features improves over the no-reconstruction variant, but remains worse than SCFAE.
This implies that a contrastive objective can introduce optimization signals that compete with the retrieval task.
In contrast, reconstruction serves as a structural constraint and keeps training driven by task gradients.
These results recommend using the full SCFAE configuration with both expansion and reconstruction for best performance.

\begin{table}[t]
    \centering
    \caption{Ablation of reconstruction metric on ActivityNet. Cosine similarity is more stable for heterogeneous encoder features, while MSE remains competitive.}
    \label{tab:mse_cos_ablation}
    \setlength{\tabcolsep}{5pt}
    \resizebox{0.78\columnwidth}{!}{
    \begin{tabular}{lccc}
    \toprule
    Method & mAP@10 & mAP@50 & mAP@100 \\
    \midrule
    MSE, 128--128 & 0.358 & 0.338 & 0.315 \\
    Cos, 128--128 & 0.363 & 0.343 & 0.321 \\
    MSE, 4096--128 & 0.365 & 0.345 & 0.323 \\
    Cos, 4096--128 & \textbf{0.367} & \textbf{0.349} & \textbf{0.326} \\
    \bottomrule
    \end{tabular}
    }
\end{table}

\textbf{Reconstruction metric.}
Tab.~\ref{tab:mse_cos_ablation} compares MSE and cosine similarity for the reconstruction term. Both metrics validate the same design principle. Preserving recoverable information improves retrieval regardless of the distance function. Cosine similarity is more suitable for heterogeneous encoders whose feature scales differ across modalities. Directional consistency is a more reliable reconstruction signal than matching absolute magnitudes.
We recommend cosine similarity as the default reconstruction metric.

\subsection{Parameter Sensitivity}
\label{sec:sensitivity}

We evaluate how the expansion factor $n$ and boundary ratio $b$ affect the model. We use the CMU-MOSEI dataset. We report accuracy on the audio-visual subset where the dominant text modality is missing. This setting tests if the model learns robust features without relying on the strongest signal. Tab.~\ref{tab:sensitivity} shows the results.

\begin{table}[t!]
    \centering
    \caption{Parameter sensitivity on CMU-MOSEI (Testing on $\{a, v\}$ subset). We report accuracy (\%) for different combinations of expansion factor $n$ and boundary ratio $b$. The default setting ($n=2, b=0.5$) provides the best balance.}
    \label{tab:sensitivity}
    \setlength{\tabcolsep}{6mm}
    \resizebox{0.8\columnwidth}{!}{
    \begin{tabular}{c|ccc}
    \toprule
    \multirow{2}{*}{Expansion $n$} & \multicolumn{3}{c}{Boundary Ratio $b$} \\
    \cmidrule(lr){2-4}
     & 0.25 & 0.5 & 0.75 \\
    \midrule
    1 & 69.1 & 70.1 & 69.5 \\
    2 & 71.2 & \textbf{72.0} & 71.5 \\
    4 & 71.4 & 72.1 & 71.6 \\
    \bottomrule
    \end{tabular}
    }
    \vskip -1em
\end{table}

\textbf{Expansion factor.} Increasing $n$ improves performance up to a plateau. A small feature space cannot support effective separation. Once $n\ge 2$, performance stabilizes. This suggests that sufficient representational capacity is the key requirement.
We recommend setting $n=2$ as a safe default that balances capacity and efficiency.

\textbf{Boundary ratio.} The ratio $b$ controls the split between shared and specific space. A small $b$ reduces shared space and hurts task-related information. A large $b$ reduces specific space and weakens noise filtering. Overall, SCFAE is not sensitive to $b$ once the expansion space is large enough. The default $b=0.5$ works well across settings.

\begin{table}[t]
    \centering
    \caption{Sensitivity to reconstruction weight $\lambda$ on CMU-MOSEI. We report accuracy (\%).}
    \label{tab:lambda_sensitivity}
    \setlength{\tabcolsep}{10pt}
    \resizebox{0.8\columnwidth}{!}{
    \begin{tabular}{ccccc}
    \toprule
    $\lambda$ & $\{a\}$ & $\{v\}$ & $\{a,v\}$ & $\{a,t,v\}$ \\
    \midrule
    0.1 & 68.2 & 69.1 & 70.8 & 86.9 \\
    0.5 & 68.9 & 69.8 & 71.5 & 87.5 \\
    1.0 & \textbf{69.3} & \textbf{70.4} & \textbf{72.0} & \textbf{87.8} \\
    2.0 & 68.7 & 70.1 & 71.6 & 87.6 \\
    5.0 & 67.8 & 69.3 & 71.1 & 87.2 \\
    \bottomrule
    \end{tabular}
    }
    \vskip -1.5em
\end{table}

\textbf{Reconstruction weight.}
Tab.~\ref{tab:lambda_sensitivity} shows that SCFAE is not highly fragile to $\lambda$. The best result is near the default $\lambda=1.0$, but performance remains stable over the intermediate range from 0.5 to 2.0. When $\lambda$ is too small, the reconstruction constraint becomes weak and the model moves toward ordinary task-driven fusion. When $\lambda$ is too large, reconstruction is over-emphasized and discriminative information is mildly suppressed.
The results suggest $\lambda=1.0$ as a robust default without task-specific tuning.

\subsection{Visualization and Gradient Analysis}
\label{sec:vis}

We visualize training dynamics in Fig.~\ref{fig:grads} to test our claims about gradient flow and feature organization.

\textbf{SCFAE removes optimization conflict.} Fig.~\ref{fig:grads}(A) compares the task gradient with the auxiliary gradient at the encoder. SCFAE stays near zero similarity most of the time, so the two objectives rarely oppose each other. Several baselines show a broad spread and many negative values. These negatives mean the auxiliary loss can push the encoder against the task, which weakens learning.

\textbf{Task gradients reach the encoder cleanly.} Fig.~\ref{fig:grads}(B) tracks how much task gradient reaches the encoder during training. Coupled fusion can quickly shrink this signal, which matches the bottleneck effect. Competing losses can also introduce oscillations. SCFAE keeps the gradient strong and steady, so the task can keep training the encoder.

\textbf{Features separate naturally.} Fig.~\ref{fig:grads}(C) shows a t-SNE view of the learned features. Shared components from different modalities cluster together, while modality-specific components separate into distinct subspaces. This behavior comes from the reconstruction constraint, so we do not need extra contrastive losses to enforce it.

\vspace{-0.09cm}
\section{Conclusion}
This paper proposes SCFAE, a self-consistent framework organizing multimodal fusion through dual forces: task loss as drift and reconstruction loss as diffusion. It provides a design criterion: the fusion module must function as both an information coupler and a gradient distributor, with complementary objectives operating through distinct feature subspaces. Following this criterion prevents gradient interference and preserves encoder robustness under missing modalities, at the cost of extra computation from feature expansion.

\textbf{Limitation.} Expanding features by a factor of $n$ increases parameter count and computation quadratically with feature dimension. Future work will explore lightweight alternatives to the expansion mechanism and investigate the benefits of stacking multiple SCFAE layers.

\section*{Impact Statement}
This paper presents work whose goal is to advance the field of Machine Learning. There are many potential societal consequences of our work, none which we feel must be specifically highlighted here.

\section*{Acknowledgements}

I would like to express my sincere gratitude to Prof. Jing Wang for the continuous support, patience, and trust throughout the long development of this work. 

This paper has gone through nearly two years of rejections, revisions, and rethinking. During this process, we repeatedly improved the method, refined the experiments, and tried to better explain why the model works. One important lesson I learned is that heuristic algorithms, even when they are effective, are not always convincing unless their assumptions and mechanisms are made clear. Another lesson is that physical analogies should not be used only as intuitive motivation. They need stricter constraints, clearer definitions, and more careful validation before they can serve as a solid foundation for machine learning methods. This work therefore reflects not only a technical attempt at multimodal fusion, but also a process of learning how to turn intuition into a more disciplined research argument.


\bibliography{example_paper}

@inproceedings{kawaguchi2023information,
  title={How Does Information Bottleneck Help Deep Learning?},
  author={Kawaguchi, K. and Deng, Z. and Ji, X. and Huang, J.},
  booktitle={Proceedings of the 40th International Conference on Machine Learning},
  year={2023},
  url={https://proceedings.mlr.press/v202/kawaguchi23a/kawaguchi23a.pdf},
  note={Accessed: 2025-09-25}
}

@article{Schuss2001DerivationPNP,
  title   = {Derivation of Poisson and Nernst--Planck Equations in a Bath and Channel from a Molecular Model},
  author  = {Schuss, Zeev and Nadler, Boaz and Eisenberg, Robert S.},
  journal = {Physical Review E},
  year    = {2001},
  volume  = {64},
  number  = {3},
  pages   = {036116},
  doi     = {10.1103/PhysRevE.64.036116}
}

@Article{ContrastiveLearning,
AUTHOR = {Jaiswal, Ashish and Babu, Ashwin Ramesh and Zadeh, Mohammad Zaki and Banerjee, Debapriya and Makedon, Fillia},
TITLE = {A Survey on Contrastive Self-Supervised Learning},
JOURNAL = {Technologies},
VOLUME = {9},
YEAR = {2021},
NUMBER = {1},
ARTICLE-NUMBER = {2}
}

@INPROCEEDINGS{MLA,
  author={Zhang, Xiaohui and Yoon, Jaehong and Bansal, Mohit and Yao, Huaxiu},
  booktitle={IEEE/CVF Conference on Computer Vision and Pattern Recognition (CVPR)},
  title={Multimodal Representation Learning by Alternating Unimodal Adaptation}, 
  year={2024},
  pages={27446-27456}
}

@article{Missing,
  title={Multi-Modal Learning with Missing Modality via Shared-Specific Feature Modelling},
  author={Hu Wang and Yuanhong Chen and Congbo Ma and Jodie Avery and Louise Hull and G. Carneiro},
  journal={IEEE/CVF Conference on Computer Vision and Pattern Recognition (CVPR)},
  year={2023},
  pages={15878-15887}
}

@article{MDF,
  title={MDF-FND: A dynamic fusion model for multimodal fake news detection},
  author={Hongzhen Lv and Wenzhong Yang and Yabo Yin and Fuyuan Wei and Jiaren Peng and Haokun Geng},
  journal={Knowl. Based Syst.},
  year={2025},
  volume={317},
  pages={113417}
}

@InProceedings{QMF,
  title = 	 {Provable Dynamic Fusion for Low-Quality Multimodal Data},
  author =       {Zhang, Qingyang and Wu, Haitao and Zhang, Changqing and Hu, Qinghua and Fu, Huazhu and Zhou, Joey Tianyi and Peng, Xi},
  booktitle = 	 {International Conference on Machine Learning (ICML)},
  pages = 	 {41753--41769},
  year = 	 {2023},
  volume = 	 {202}
}

@inproceedings{CLIP,
  title={Learning Transferable Visual Models From Natural Language Supervision},
  author={Alec Radford and Jong Wook Kim and Chris Hallacy and Aditya Ramesh and Gabriel Goh and Sandhini Agarwal and Girish Sastry and Amanda Askell and Pamela Mishkin and Jack Clark and Gretchen Krueger and Ilya Sutskever},
  booktitle={International Conference on Machine Learning (ICML)},
  pages={8748-8763},
  year={2021}
}

@inproceedings{METER,
  title={An Empirical Study of Training End-to-End Vision-and-Language Transformers},
  author={Zi-Yi Dou and Yichong Xu and Zhe Gan and Jianfeng Wang and Shuohang Wang and Lijuan Wang and Chenguang Zhu and Nanyun Peng and Zicheng Liu and Michael Zeng},
  booktitle={IEEE/CVF Conference on Computer Vision and Pattern Recognition (CVPR)},
  pages={18145-18155},  
  year={2021}
}

@inproceedings{MAP-IVR,
  title={Activity Image-to-Video Retrieval by Disentangling Appearance and Motion},
  author={Liu Liu and Jiangtong Li and Li Niu and Ruicong Xu and Liqing Zhang},
  booktitle={Association for the Advancement of Artificial Intelligence (AAAI)},
  page={2145-2153},
  year={2021}
}

@inproceedings{APIVR, 
  title={A Proposal-Based Approach for Activity Image-to-Video Retrieval}, 
  author={Xu, Ruicong and Niu, Li and Zhang, Jianfu and Zhang, Liqing}, 
  booktitle={Association for the Advancement of Artificial Intelligence (AAAI)}, 
  pages={12524-12531}, 
  year={2020}
}

@article{CoupledMamba2024,
  title={Coupled mamba: Enhanced multimodal fusion with coupled state space model},
  author={Li, Wenbing and Zhou, Hang and Yu, Junqing and Song, Zikai and Yang, Wei},
  journal={Advances in Neural Information Processing Systems (NeurIPS)},
  volume={37},
  pages={59808--59832},
  year={2024}
}

@article{SALM2025,
  title={Multimodal Emotion Recognition via the Fusion of Mamba and Liquid Neural Networks with Cross-Modal Alignment},
  author={Chen, Guoming and Liao, Yuting and Zhang, Dong and Yang, Weikang and Mai, Ziying and Xu, Chenying},
  journal={Electronics},
  volume={14},
  number={18},
  pages={3638},
  year={2025},
  publisher={MDPI}
}

@article{AVoiD-DF,
  title={AVoiD-DF: Audio-Visual Joint Learning for Detecting Deepfake}, 
  author={Yang, Wenyuan and Zhou, Xiaoyu and Chen, Zhikai and Guo, Bofei and Ba, Zhongjie and Xia, Zhihua and Cao, Xiaochun and Ren, Kui},
  journal={IEEE Transactions on Information Forensics and Security}, 
  pages={2015-2029},
  year={2023}
}

@inproceedings{MISA,
  title={MISA: Modality-Invariant and -Specific Representations for Multimodal Sentiment Analysis},
  author={Devamanyu Hazarika and Roger Zimmermann and Soujanya Poria},
  booktitle={ACM International Conference on Multimedia (ACM MM)},
  pages={1122-1131},
  year={2020}
}

@article{UAVM,
  title={UAVM: Towards Unifying Audio and Visual Models},
  author={Yuan Gong and Alexander H. Liu and Andrew Rouditchenko and James R. Glass},
  journal={IEEE Signal Processing Letters},
  pages={2437-2441},
  year={2022}
}

@inproceedings{DrFuse, 
  title={DrFuse: Learning Disentangled Representation for Clinical Multi-Modal Fusion with Missing Modality and Modal Inconsistency}, 
  author={Yao, Wenfang and Yin, Kejing and Cheung, William K. and Liu, Jia and Qin, Jing}, 
  booktitle={Association for the Advancement of Artificial Intelligence (AAAI)}, 
  pages={16416-16424}, 
  year={2024}
}

@inproceedings{MBT,
  title={Attention Bottlenecks for Multimodal Fusion},
  author={Arsha Nagrani and Shan Yang and Anurag Arnab and Aren Jansen and Cordelia Schmid and Chen Sun},
  booktitle={Conference and Workshop on Neural Information Processing Systems (NeurIPS)},
  pages={14200-14213},
  year={2021}
}

@inproceedings{Perceiver,
  title={Perceiver: General Perception with Iterative Attention},
  author={Andrew Jaegle and Felix Gimeno and Andrew Brock and Andrew Zisserman and Oriol Vinyals and João Carreira},
  booktitle={International Conference on Machine Learning (ICML)},
  page={4651-4664},
  year={2021}
}

@inproceedings{AnchorDML,
  title     = {Anchor-aware Deep Metric Learning for Audio-visual Retrieval},
  author    = {Zeng, Donghuo and Wang, Yanan and Ikeda, Kazushi and Yu, Yi},
  booktitle = {Proceedings of the International Conference on Multimedia Retrieval (ICMR)},
  publisher = {ACM},
  year      = {2024},
  pages     = {211--219}
}

@article{DI-VTR,
  title   = {DI-VTR: Dual Inter-modal Interaction Model for Video-Text Retrieval},
  author  = {Guo, Jie and Wang, Mengying and Wang, Wenwei and Zhou, Yan and Song, Bin},
  journal = {Journal of Information and Intelligence},
  year    = {2024},
  volume  = {2},
  number  = {5},
  pages   = {388--403}
}

@article{CMKT,
  title   = {Bridging asymmetry between image and video: Cross-modality knowledge transfer based on learning from video},
  author  = {Zhou, Bingxin and Zhou, Jianghao and Chen, Zhongming and Jiang, Hanyuan and Zou, Lei},
  journal = {Expert Systems with Applications (ESWA)},
  year    = {2025},
  volume  = {264},
  pages   = {125873}
}

@inproceedings{LossEqualToMaxMI,
  title={A Unifying Mutual Information View of Metric Learning: Cross-Entropy vs. Pairwise Losses},
  author={Malik Boudiaf and Jérôme Rony and Imtiaz Masud Ziko and Eric Granger and Marco Pedersoli and Pablo Piantanida and Ismail Ben Ayed},
  booktitle={European Conference on Computer Vision (ECCV)},
  pages = {548–564},
  year={2020}
}

@inproceedings{ContrLoss,
  title={Understanding the Behaviour of Contrastive Loss},
  author={Feng Wang and Huaping Liu},
  booktitle={IEEE/CVF Conference on Computer Vision and Pattern Recognition (CVPR)},
  pages={2495-2504},
  year={2020}
}

@inproceedings{GAP,
 title = {Mind the Gap: Understanding the Modality Gap in Multi-modal Contrastive Representation Learning},
 author = {Liang, Victor Weixin and Zhang, Yuhui and Kwon, Yongchan and Yeung, Serena and Zou, James Y},
 booktitle = {Advances in Neural Information Processing Systems (NeurIPS)},
 pages = {17612--17625},
 year = {2022}
}

@inproceedings{GradRelated,
author = {Du, Chenzhuang and Teng, Jiaye and Li, Tingle and Liu, Yichen and Yuan, Tianyuan and Wang, Yue and Yuan, Yang and Zhao, Hang},
title = {On uni-modal feature learning in supervised multi-modal learning},
year = {2023},
booktitle = {International Conference on Machine Learning (ICML)},
articleno = {345},
numpages = {25}
}

@article{PNP,
  title={Entropy method for generalized Poisson–Nernst–Planck equations},
  author={José Rodrigo González Granada and Victor A. Kovtunenko},
  journal={Analysis and Mathematical Physics},
  pages={603-619},
  year={2018}
}

@INPROCEEDINGS{MMEntro,
  title={Understanding and Constructing Latent Modality Structures in Multi-Modal Representation Learning}, 
  author={Jiang, Qian and Chen, Changyou and Zhao, Han and Chen, Liqun and Ping, Qing and Tran, Son Dinh and Xu, Yi and Zeng, Belinda and Chilimbi, Trishul},
  booktitle={IEEE/CVF Conference on Computer Vision and Pattern Recognition (CVPR)}, 
  pages={7661-7671},
  year={2023}
}

@INPROCEEDINGS{ActivityNet,
  title={ActivityNet: A large-scale video benchmark for human activity understanding}, 
  author={Heilbron, Fabian Caba and Escorcia, Victor and Ghanem, Bernard and Niebles, Juan Carlos},
  booktitle={IEEE Conference on Computer Vision and Pattern Recognition (CVPR)}, 
  year={2015},
  pages={961-970}
}

@inproceedings{FAV,
 title = {FakeAVCeleb: A Novel Audio-Video Multimodal Deepfake Dataset},
 author = {Khalid, Hasam and Tariq, Shahroz and Kim, Minha and Woo, Simon},
 booktitle = {Proceedings of the Neural Information Processing Systems Track on Datasets and Benchmarks},
 year = {2021}
}

@INPROCEEDINGS{ResNet,
  title={Deep Residual Learning for Image Recognition}, 
  author={He, Kaiming and Zhang, Xiangyu and Ren, Shaoqing and Sun, Jian},
  booktitle={IEEE Conference on Computer Vision and Pattern Recognition (CVPR)}, 
  pages={770-778},
  year={2016}
}

@inproceedings{VAE,
  title = {Auto-Encoding Variational Bayes},
  author = {Kingma, Diederik P. and Welling, Max},
  booktitle = {International Conference on Learning Representations (ICLR)},
  year = {2014}
}

@InProceedings{R2+1D,
title = {A Closer Look at Spatiotemporal Convolutions for Action Recognition},
author = {Tran, Du and Wang, Heng and Torresani, Lorenzo and Ray, Jamie and LeCun, Yann and Paluri, Manohar},
booktitle = {IEEE Conference on Computer Vision and Pattern Recognition (CVPR)},
pages = {6450-6459},
year = {2018}
}

@ARTICLE{LSTM,
  title={Long Short-Term Memory}, 
  author={Hochreiter, Sepp and Schmidhuber, Jürgen},
  year={1997},
  journal={Neural Computation}, 
  pages={1735-1780}
}

@inproceedings{Emo,
  title={Emotions Don't Lie: An Audio-Visual Deepfake Detection Method using Affective Cues},
  author={Trisha Mittal and Uttaran Bhattacharya and Rohan Chandra and Aniket Bera and Dinesh Manocha},
  booktitle={ACM International Conference on Multimedia (ACM MM)},
  year={2020},
  pages = {2823–2832}
}

@inproceedings{MDS,
  title={Not made for each other- Audio-Visual Dissonance-based Deepfake Detection and Localization},
  author={Komal Chugh and Parul Gupta and Abhinav Dhall and Ramanathan Subramanian},
  booktitle={ACM International Conference on Multimedia (ACM MM)},
  year={2020},
  pages = {439–447}
}

@article{VFD,
title = {Voice-Face Homogeneity Tells Deepfake},
author = {Cheng, Harry and Guo, Yangyang and Wang, Tianyi and Li, Qi and Chang, Xiaojun and Nie, Liqiang},
journal = {ACM Transactions on Multimedia Computing, Communications, and Applications},
year = {2023}
}

@INPROCEEDINGS{DenseNet,
  author={Huang, Gao and Liu, Zhuang and Van Der Maaten, Laurens and Weinberger, Kilian Q.},
  booktitle={IEEE Conference on Computer Vision and Pattern Recognition (CVPR)}, 
  title={Densely Connected Convolutional Networks}, 
  year={2017},
  pages={2261-2269}
}

@misc{mmres,
      title={Progressive Fusion for Multimodal Integration}, 
      author={Shiv Shankar and Laure Thompson and Madalina Fiterau},
      year={2022},
      eprint={2209.00302},
      archivePrefix={arXiv},
      primaryClass={cs.LG}
}

@article{Baltrusaitis2019Survey,
  author    = {Tadas Baltru{\v{s}}aitis and Chaitanya Ahuja and Louis{-}Philippe Morency},
  title     = {Multimodal Machine Learning: A Survey and Taxonomy},
  journal   = {IEEE Transactions on Pattern Analysis and Machine Intelligence (TPAMI)},
  volume    = {41},
  number    = {2},
  pages     = {423--443},
  year      = {2019}
}

@inproceedings{Lu2019ViLBERT,
  author    = {Jiasen Lu and Dhruv Batra and Devi Parikh and Stefan Lee},
  title     = {ViLBERT: Pretraining Task-Agnostic Visiolinguistic Representations for Vision-and-Language Tasks},
  booktitle = {Advances in Neural Information Processing Systems (NeurIPS)},
  year      = {2019},
  volume    = {32}
}

@article{Zhao2024CSUR,
  author    = {Fei Zhao and Chengcui Zhang and Baocheng Geng},
  title     = {Deep Multimodal Data Fusion},
  journal   = {ACM Computing Surveys (ACM CSUR)},
  volume    = {56},
  number    = {9},
  pages     = {1--36},
  year      = {2024}
}

@inproceedings{Kendall2018Uncertainty,
  author    = {Alex Kendall and Yarin Gal and Roberto Cipolla},
  title     = {Multi-Task Learning Using Uncertainty to Weigh Losses for Scene Geometry and Semantics},
  booktitle = {Proceedings of the IEEE/CVF Conference on Computer Vision and Pattern Recognition (CVPR)},
  pages     = {7482--7491},
  year      = {2018}
}

@inproceedings{Chen2018GradNorm,
  author    = {Zhao Chen and Vijay Badrinarayanan and Chen{-}Yu Lee and Andrew Rabinovich},
  title     = {GradNorm: Gradient Normalization for Adaptive Loss Balancing in Deep Multitask Networks},
  booktitle = {Proceedings of the International Conference on Machine Learning (ICML)},
  volume    = {80},
  pages     = {794--803},
  year      = {2018}
}

@inproceedings{MCTN,
  title={Found in translation: Learning robust joint representations by cyclic translations between modalities},
  author={Pham, Hai and Liang, Paul Pu and Manzini, Thomas and Morency, Louis-Philippe and P{\'o}czos, Barnab{\'a}s},
  booktitle={Proceedings of the AAAI conference on artificial intelligence},
  volume={33},
  number={01},
  pages={6892--6899},
  year={2019}
}

@inproceedings{MMIN,
  title={Missing modality imagination network for emotion recognition with uncertain missing modalities},
  author={Zhao, Jinming and Li, Ruichen and Jin, Qin},
  booktitle={Proceedings of the 59th Annual Meeting of the Association for Computational Linguistics and the 11th International Joint Conference on Natural Language Processing (Volume 1: Long Papers)},
  pages={2608--2618},
  year={2021}
}

@article{GCNet,
  title={Gcnet: Graph completion network for incomplete multimodal learning in conversation},
  author={Lian, Zheng and Chen, Lan and Sun, Licai and Liu, Bin and Tao, Jianhua},
  journal={IEEE Transactions on pattern analysis and machine intelligence},
  volume={45},
  number={7},
  pages={8419--8432},
  year={2023},
  publisher={IEEE}
}

@article{IMDer,
  title={Incomplete multimodality-diffused emotion recognition},
  author={Wang, Yuanzhi and Li, Yong and Cui, Zhen},
  journal={Advances in Neural Information Processing Systems},
  volume={36},
  pages={17117--17128},
  year={2023}
}

@inproceedings{DiCMoR,
  title={Distribution-consistent modal recovering for incomplete multimodal learning},
  author={Wang, Yuanzhi and Cui, Zhen and Li, Yong},
  booktitle={Proceedings of the IEEE/CVF International Conference on Computer Vision},
  pages={22025--22034},
  year={2023}
}

@inproceedings{MoMKE,
  title={Leveraging knowledge of modality experts for incomplete multimodal learning},
  author={Xu, Wenxin and Jiang, Hexin and Liang, Xuefeng},
  booktitle={Proceedings of the 32nd ACM International Conference on Multimedia},
  pages={438--446},
  year={2024}
}

@inproceedings{EUAR,
  title={Enhanced Experts with Uncertainty-Aware Routing for Multimodal Sentiment Analysis},
  author={Gao, Zixian and Hu, Disen and Jiang, Xun and Lu, Huimin and Shen, Heng Tao and Xu, Xing},
  booktitle={Proceedings of the 32nd ACM International Conference on Multimedia},
  pages={9650--9659},
  year={2024}
}

@article{MCULoRA,
  title={A Robust Incomplete Multimodal Low-Rank Adaptation Approach for Emotion Recognition},
  author={Zhao, Xinkui and Shu, Jinsong and Wu, Yangyang and Cheng, Guanjie and Liu, Zihe and Wang, Naibo and Deng, Shuiguang and Xie, Zhongle and Yin, Jianwei},
  journal={arXiv preprint arXiv:2507.11202},
  year={2025}
}

@inproceedings{MOSEI,
    title = "Multimodal Language Analysis in the Wild: {CMU}-{MOSEI} Dataset and Interpretable Dynamic Fusion Graph",
    author = "Bagher Zadeh, AmirAli  and
      Liang, Paul Pu  and
      Poria, Soujanya  and
      Cambria, Erik  and
      Morency, Louis-Philippe",
    booktitle = "Proceedings of the 56th Annual Meeting of the Association for Computational Linguistics (Volume 1: Long Papers)",
    month = jul,
    year = "2018",
    pages = "2236--2246"
}

@inproceedings{HARD1,
  title={What makes training multi-modal classification networks hard?},
  author={Wang, Weiyao and Tran, Du and Feiszli, Matt},
  booktitle={Proceedings of the IEEE/CVF conference on computer vision and pattern recognition},
  pages={12695--12705},
  year={2020}
}

@inproceedings{wang2025audio,
  title={Audio-Visual Asynchrony Mitigation: Cross-Modal Alignment and Feature Reconstruction for Deepfake Detection},
  author={Wang, Yan and Sun, Qindong and Rong, Dongzhu},
  booktitle={Proceedings of the 33rd ACM International Conference on Multimedia},
  pages={11542--11551},
  year={2025}
}

@inproceedings{MACBDF,
  title={Multiscale Adaptive Conflict-Balancing Model For Multimedia Deepfake Detection},
  author={Xiong, Zihan and Wu, Xiaohua and Chen, Lei and Lou, Fangqi},
  booktitle={Proceedings of the 2025 International Conference on Multimedia Retrieval (ICMR)},
  pages={1598--1606},
  year={2025}
}

@article{WangHuang2024,
  title={Audio--visual deepfake detection using articulatory representation learning},
  author={Wang, Yujia and Huang, Hua},
  journal={Computer Vision and Image Understanding},
  volume={248},
  pages={104133},
  year={2024},
  publisher={Elsevier}
}

@inproceedings{AdaMMS2025,
  title={Adamms: Model merging for heterogeneous multimodal large language models with unsupervised coefficient optimization},
  author={Du, Yiyang and Wang, Xiaochen and Chen, Chi and Ye, Jiabo and Wang, Yiru and Li, Peng and Yan, Ming and Zhang, Ji and Huang, Fei and Sui, Zhifang and others},
  booktitle={Proceedings of the Computer Vision and Pattern Recognition Conference (CVPR)},
  pages={9413--9422},
  year={2025}
}

@inproceedings{DLF2025,
  title={DLF: Disentangled-language-focused multimodal sentiment analysis},
  author={Wang, Pan and Zhou, Qiang and Wu, Yawen and Chen, Tianlong and Hu, Jingtong},
  booktitle={Proceedings of the AAAI Conference on Artificial Intelligence},
  volume={39},
  number={20},
  pages={21180--21188},
  year={2025}
}

@inproceedings{girdhar2023imagebind,
  title={Imagebind: One embedding space to bind them all},
  author={Girdhar, Rohit and El-Nouby, Alaaeldin and Liu, Zhuang and Singh, Mannat and Alwala, Kalyan Vasudev and Joulin, Armand and Misra, Ishan},
  booktitle={Proceedings of the IEEE/CVF conference on computer vision and pattern recognition},
  pages={15180--15190},
  year={2023}
}

@misc{simeoni2025dinov3,
  title={{DINOv3}},
  author={Sim{\'e}oni, Oriane and Vo, Huy V. and Seitzer, Maximilian and Baldassarre, Federico and Oquab, Maxime and Jose, Cijo and Khalidov, Vasil and Szafraniec, Marc and Yi, Seungeun and Ramamonjisoa, Micha{\"e}l and Massa, Francisco and Haziza, Daniel and Wehrstedt, Luca and Wang, Jianyuan and Darcet, Timoth{\'e}e and Moutakanni, Th{\'e}o and Sentana, Leonel and Roberts, Claire and Vedaldi, Andrea and Tolan, Jamie and Brandt, John and Couprie, Camille and Mairal, Julien and J{\'e}gou, Herv{\'e} and Labatut, Patrick and Bojanowski, Piotr},
  year={2025},
  eprint={2508.10104},
  archivePrefix={arXiv},
  primaryClass={cs.CV},
  url={https://arxiv.org/abs/2508.10104},
}

@article{huang2022masked,
  title={Masked autoencoders that listen},
  author={Huang, Po-Yao and Xu, Hu and Li, Juncheng and Baevski, Alexei and Auli, Michael and Galuba, Wojciech and Metze, Florian and Feichtenhofer, Christoph},
  journal={Advances in Neural Information Processing Systems (NeurIPS)},
  volume={35},
  pages={28708--28720},
  year={2022}
}

@inproceedings{wang2023videomae,
  title={Videomae v2: Scaling video masked autoencoders with dual masking},
  author={Wang, Limin and Huang, Bingkun and Zhao, Zhiyu and Tong, Zhan and He, Yinan and Wang, Yi and Wang, Yali and Qiao, Yu},
  booktitle={Proceedings of the IEEE/CVF conference on computer vision and pattern recognition},
  pages={14549--14560},
  year={2023}
}

@article{chen2022wavlm,
  title={Wavlm: Large-scale self-supervised pre-training for full stack speech processing},
  author={Chen, Sanyuan and Wang, Chengyi and Chen, Zhengyang and Wu, Yu and Liu, Shujie and Chen, Zhuo and Li, Jinyu and Kanda, Naoyuki and Yoshioka, Takuya and Xiao, Xiong and others},
  journal={IEEE Journal of Selected Topics in Signal Processing},
  volume={16},
  number={6},
  pages={1505--1518},
  year={2022},
  publisher={IEEE}
}
\bibliographystyle{icml2026}


\newpage
\appendix
\onecolumn
\section{Additional Experiments}
\label{append:expr}
This section provides additional details on the implementation, introduction of the dataset, and analysis of the experimental results.

\begin{table}[t!]
\scriptsize
\centering
\caption{Parameter count and computational complexity for SCFAE (Linear Mapping) matrices per modality}
\begin{tabular}{lll|rr}
\hline
\multicolumn{3}{c|}{\textbf{Modality } $m$} & \textbf{Parameters} & \textbf{FLOPs} \\
\hline
\multirow{6}{*}{\textbf{Matrix}}
& \multicolumn{2}{l|}{\textbf{SwiGLU} (pre)}
& $8(l^{(m)})^2 + \frac{19}{3}l^{(m)}$
& $16(l^{(m)})^2 + \frac{19}{3}l^{(m)}$ \\

& \multicolumn{2}{l|}{$\mathbf{P}_{\mathrm{expand}}^{(m)}$}
& $n \cdot l^{(m)} \times l^{(m)} + n \cdot l^{(m)}$
& $2n \cdot (l^{(m)})^2 + n \cdot l^{(m)}$ \\

& \multicolumn{2}{l|}{$\mathbf{P}_{\mathrm{shared}}^{(m)}$}
& $b^{(m)} \times l^{*} + b^{(m)}$
& $2 b^{(m)} l^{*} + b^{(m)}$ \\

& \multicolumn{2}{l|}{$\mathbf{P}_{\mathrm{specific}}^{(m)}$}
& $(n l^{(m)} - b^{(m)}) \times l^{(m)} + (n l^{(m)} - b^{(m)})$
& $2 (n l^{(m)} - b^{(m)}) l^{(m)} + (n l^{(m)} - b^{(m)})$ \\

& \multicolumn{2}{l|}{$\mathbf{P}_{\mathrm{recon}}^{(m)}$}
& $l^{(m)} \times (l^{(m)} + l^{*}) + l^{(m)}$
& $2 l^{(m)} (l^{(m)} + l^{*}) + l^{(m)}$ \\

& \multicolumn{2}{l|}{\textbf{SwiGLU} (post)}
& $8(l^{(m)}+l^{*})^2 + \frac{19}{3}(l^{(m)}+l^{*})$
& $16(l^{(m)}+l^{*})^2 + \frac{19}{3}(l^{(m)}+l^{*})$ \\
\hline
\end{tabular}
\label{tab:parameters_flops}
\end{table}

\textbf{Structure.} As shown in~\ref{tab:parameters_flops}. Any matrix in $\theta^\text{SCFAE}$ is a standard linear layer, and then it goes through GeLU. During the input and output processes, in order to balance the parameter quantity, there is a set of SwiGLU with an intermediate dimension of $8/3d$. For FakeAVCeleb~\citep{FAV}, we use both R(2+1)D~\citep{R2+1D}. SCFAE module contains 24.01 M params and 0.048 GFLOPs (512d for audio, 512d for visual).  For ActivityNet~\citep{ActivityNet}, we use features extracted by~\citet{APIVR}, and 128-128d groups using a simple linear map 4096d video feature to 128d. On CMU-MOSEI~\citep{MOSEI}, our feature extraction structure is exactly the same as that of MCULoRA~\citep{MCULoRA}.

\textbf{Implementation Details.} We use AdamW optimizer with betas = $(0.9, 0.999)$. The initial learning rate is set at 2.5e-5. It is linearly warmed up to 5e-4 after 20\% of epochs, and then decays using a cosine function. We trained for 10 epochs on FakeAVCeleb and ActivityNet, and for 125 epochs on CMU-MOSEI. The weight of reconstruction loss $\lambda = 1$.

\subsection{Codes.} Given the availability of robust and comprehensive implementations for certain baselines (e.g., GCNet~\citep{GCNet}), we have refrained from redundant reimplementation. We provide only the code developed specifically for our reproduction, and direct readers to the official repositories associated with the cited works for the remaining components. This approach serves to honor the original contributors' efforts while strictly adhering to ICML anonymity protocols, which preclude the inclusion of potentially deanonymizing links in the supplementary material.

\subsection{Datasets and Additional Analysis.} 

\textbf{FakeAVCeleb.} We evaluate deepfake detection on FakeAVCeleb~\citep{FAV}, a highly imbalanced benchmark (overall real:fake $\approx$ 1:39; video 1:19; audio 1:1). To ensure fair comparison, we adopt the feature extractors used in the original methods: MISA~\citep{MISA} with sLSTM~\citep{LSTM}, and R(2+1)D-34~\citep{R2+1D} for DrFuse~\citep{DrFuse}, Perceiver~\citep{Perceiver}, our concat baseline, and SCFAE backbones. For AVoiD-DF~\citep{AVoiD-DF}, VFD~\citep{VFD}, Emo-Foren~\citep{Emo}, and MDS~\citep{MDS}, we report the results published in \citet{AVoiD-DF} under the same split. The rest of method we report the results published in original paper.

\textbf{ActivityNet.} We evaluate image--video retrieval on ActivityNet under two input configurations: unequal-length (128d-4096d, 128d image, 4096d video, following~\citet{APIVR}) and equal-length (128d-128d, mapping video feature to 128d additionally). Baselines include alignment method: CMMP~\citep{CLIP}, METER~\citep{METER}; disentanglement method (shared only): APIVR~\citep{APIVR}, MAP-IVR~\citep{MAP-IVR}, DrFuse~\citep{DrFuse}, MISA~\citep{MISA}; and structure-focused methods: Perceiver~\citep{Perceiver}, DI-VTR~\citep{DI-VTR}, and guided metric learning based: AADML~\citep{AnchorDML}, CMKT~\citep{CMKT}. Among these methods, UAVM and Perceiver rely solely on label-guided supervision; CMMP, AADML, DI-VTR, CMKT, and SCFAE use purely contrastive learning; whereas METER, MISA, DrFuse, MAP-IVR, and APIVR combine both paradigms.

\section{Additional Rebuttal Results}
\label{append:rebuttal}
This section collects the extra experiments added during rebuttal. We keep them separate from the original experimental narrative so that the new evidence can be checked independently.

\subsection{Runtime and Memory}
We profile the 512d/512d audio-visual setting on an RTX 4090 with batch size 1 and FP32. The standalone SCFAE module requires only 0.72 ms for a forward pass, and adding it to the two encoders increases total forward latency from 35.10 ms to 35.32 ms. Training shows the same pattern: the full step increases from 125.85 ms to 128.97 ms, while peak memory remains dominated by the backbone encoders.

\begin{table}[h]
\centering
\caption{Supplementary runtime profiling on RTX 4090 (batch size 1, FP32).}
\label{tab:runtime_profile}
\setlength{\tabcolsep}{4pt}
\resizebox{\textwidth}{!}{
\begin{tabular}{lccccc}
\toprule
Component & Params & Latency (ms) & Throughput (samples/s) & Incremental peak memory (MB) & Total peak memory (MB) \\
\midrule
Visual encoder only & 87.57M & 31.32 & 31.93 & 609.32 & 958.81 \\
Audio encoder only & 85.57M & 4.06 & 246.43 & 47.25 & 382.61 \\
SCFAE only & 24.14M & 0.72 & 1387.36 & 1.06 & 102.62 \\
Baseline forward & 174.45M & 35.10 & 28.49 & 609.32 & 1292.06 \\
Baseline train step & 174.45M & 125.85 & 7.95 & 5311.92 & 8019.75 \\
Full SCFAE forward & 199.12M & 35.32 & 28.31 & 609.32 & 1396.71 \\
Full SCFAE train step & 199.12M & 128.97 & 7.75 & 5309.26 & 8400.38 \\
\bottomrule
\end{tabular}
}
\end{table}

\subsection{Shared and Specific Branch Masking}
We mask branches on FakeAVCeleb to separate the roles of the shared and specific components. The shared branch carries most discriminative signal, while the full model remains better than both masked variants. This indicates that the specific branch is not redundant; it contributes complementary modality-dependent information.

\begin{table}[h]
\centering
\caption{Branch masking on FakeAVCeleb. We report AUC (\%).}
\label{tab:branch_masking}
\setlength{\tabcolsep}{10pt}
\begin{tabular}{lccc}
\toprule
Setting & AUC(A) & AUC(V) & AUC(AV) \\
\midrule
Specific-only & 96.73 & 52.41 & 95.88 \\
Shared-only & 98.84 & 54.90 & 97.96 \\
SCFAE (Full) & \textbf{99.51} & \textbf{56.40} & \textbf{98.70} \\
\bottomrule
\end{tabular}
\end{table}

\subsection{Seed Stability}
We additionally ran four trials with different random seeds. On FakeAVCeleb, the final AUC is $98.63 \pm 0.08$. On ActivityNet under the 4096--128 setting, mAP@100 is $0.3263 \pm 0.0023$. These results indicate that the reported improvements are not tied to a single seed.

\section{Gradient Backward Flow}
\label{append:grad}

Gradient backward flow refers to how gradients propagate through a neural network during backpropagation, especially in multimodal fusion settings. Understanding this flow is essential for addressing gradient conflicts and enhancing model performance.

\subsection{Definition and Explanation}

\begin{figure}[h]
    \centering
    \resizebox{\textwidth}{!}{%
        \includegraphics{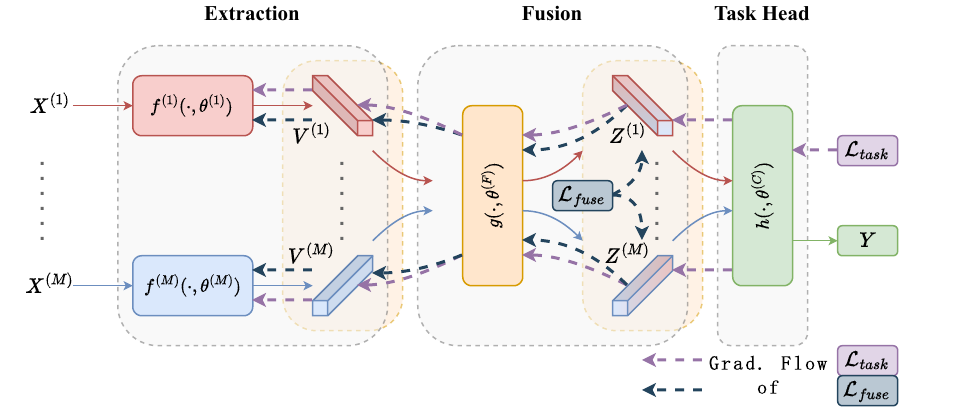}
    }
    \caption{Gradient flow diagram extended from Fig.~\ref{fig:eg}. The notation is consistent with Fig.~\ref{fig:eg}. Dark blue dotted arrows represent gradients from the fusion stage loss ($\mathcal{L}_{\text{fuse}}$), light purple dotted arrows represent gradients from the downstream task loss ($\mathcal{L}_{\text{task}}$).}
    \label{fig:gradflow}
\end{figure}

In backpropagation, gradients flow from the output back to the input, adjusting parameters based on the downstream task loss. As illustrated in Fig.~\ref{fig:gradflow}, the gradient from the downstream task loss $\mathcal{L}_{\text{task}}$ propagates through the entire network, while the gradient from the fusion stage loss $\mathcal{L}_{\text{fuse}}$ (if present) propagates from the fusion output towards the downstream task. Consequently, the feature extractor's parameters are influenced by gradients from both losses. However, gradients from $\mathcal{L}_{\text{task}}$ and $\mathcal{L}_{\text{fuse}}$ may conflict, leading to gradient interference. Balancing these conflicting gradients often requires hyperparameters to weigh the influence of different losses. For example, in variational autoencoders~\citep{VAE}, the Kullback-Leibler divergence loss promotes generalization, while the reconstruction loss focuses on accurate reconstruction. Adjusting their relative weights is crucial yet challenging.

In multimodal tasks, similar conflicts arise. Fusion-related losses, such as contrastive loss~\citep{CLIP}, aim to align features from different modalities. However, enforcing such alignment may not always be beneficial, especially when modalities carry distinct and complementary information~\citep{GAP,MMEntro}. Overemphasis on aligning modalities can lead to the loss of modality-specific information, degrading overall performance, as discussed in Sec.~\ref{sec:expr}. Introducing fusion losses directly affects feature extractors by imposing prior assumptions on the extracted features, potentially leading to suboptimal representations and increased computational cost.

\subsection{Theoretical Analysis with Residual Connections}

\begin{figure}[h]
    \centering
    \resizebox{0.75\textwidth}{!}{%
        \includegraphics{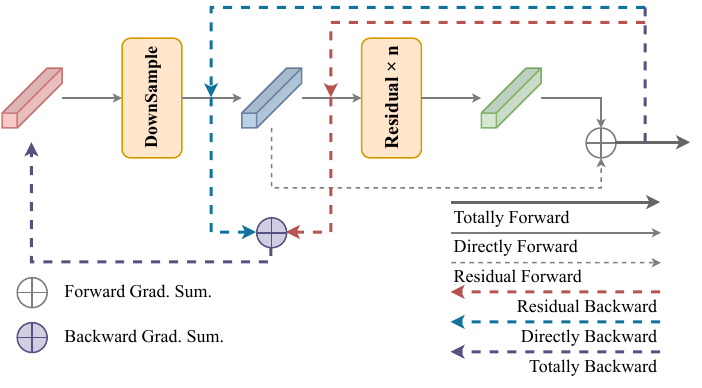}
    }
    \caption{Structure of a residual block in networks.}
    \label{fig:residual}
\end{figure}

Residual networks~\citep{ResNet} address degradation in deep networks by introducing skip connections, which alleviate gradient vanishing and exploding issues, as shown in Fig.~\ref{fig:residual}. The residual connection introduces an identity mapping that facilitates gradient flow, providing a direct path for gradients and ensuring effective training of deeper networks.

In multimodal fusion networks, similar gradient issues can arise due to deep fusion mechanisms introducing additional layers between feature extractors and downstream tasks. These layers can cause gradients to diminish or explode, leading to ineffective training of feature extractors. Applying residual connections in the fusion stage can alleviate these issues by providing direct gradient pathways~\citep{mmres, Missing}. However, residual connections require strict dimension consistency between inputs and outputs, which may not always be feasible in multimodal settings where modalities have different dimensions. Moreover, gradient interference cannot be entirely ruled out since gradients still merge in the extraction phase.

An alternative is to use concatenation-based approaches similar to DenseNet~\citep{DenseNet}, where features are stacked along the channel dimension. While this avoids some issues of residual addition, it can increase feature dimensions, causing redundancy and potential overfitting. Therefore, while residual connections can help address gradient conflicts in deep networks, they are not a complete solution for conflicts arising from conflicting loss functions in multimodal fusion. Limiting the depth of the fusion network or carefully designing fusion loss functions to align with downstream task objectives can help mitigate gradient conflicts and improve overall performance.

\end{document}